\documentclass{article}
\usepackage{microtype}
\usepackage{graphicx}
\usepackage{subfigure}
\usepackage{booktabs} 
\usepackage{url}
\usepackage{xurl}

\usepackage{hyperref}
\usepackage{url}
\usepackage{booktabs}
\usepackage{multirow}
\usepackage{makecell}
\usepackage{setspace}
\usepackage[table]{xcolor}

\usepackage[accepted]{icml2025}

\usepackage{amsmath}
\usepackage{amssymb}
\usepackage{mathtools}
\usepackage{amsthm}

\usepackage[capitalize,noabbrev]{cleveref}

\theoremstyle{plain}

\theoremstyle{definition}

\theoremstyle{remark}

\usepackage[textsize=tiny]{todonotes}

\icmltitlerunning{QRazor: Reliable and Effortless 4-bit LLM Quantization by Significant Data Razoring}

\begin{document}
\twocolumn[
\icmltitle{QRazor: Reliable and Effortless 4-bit LLM Quantization \\ by Significant Data Razoring}



\icmlsetsymbol{equal}{*}

\begin{icmlauthorlist}
\icmlauthor{Dongyoung Lee}{}
\icmlauthor{Seungkyu Choi}{}
\icmlauthor{Ik Joon Chang}{}
\end{icmlauthorlist}





\vskip 0.3in
]




\begin{abstract}
Large-scale language models (LLMs) excel in language processing tasks but face deployment challenges due to high memory and computational demands. While low-bit quantization, such as 4-bit techniques, offers a potential solution, these methods often suffer from significant accuracy loss or require considerable effort for implementation such as reordering, rotation, etc. To address these challenges, we propose QRazor, a simple yet effective quantization scheme that enables 4-bit quantization of weights, activations, and KV cache in transformer-based LLMs. QRazor operates in two stages: first, quantizing data using 8 or 16-bit integers as a basis with absolute max scaling to preserve accuracy close to full-precision models, and second, compressing the quantized data to 4-bit using our significant data razoring (SDR) technique, which retains only the four most salient bits. Without any additional requirment of fine-tuning or additional training, QRazor achieves performance similar or better compared to state-of-the-art in 4-bit quantization method, surpassing Smoothquant and QLLM by over 12 points and Quarot(RTN) by more than 2.9 points in zero-shot reasoning task accuracy on the LLaMA2-7B model. Additionally, we introduce an integer-based arithmetic unit optimized for QRazor, allowing direct low-precision operations on SDR data without decompression.
\end{abstract}

\section{Introduction}
\label{sec:intro}
Large language models (LLMs) have garnered considerable attention for their exceptional performance across various domains, including natural language processing and automated text generation. However, the extensive number of parameters in LLMs poses challenges for efficient processing. In particular, the substantial memory requirements and high computational demands can hinder their deployment in resource-constrained computing environments.

Quantization \citep{LLM.int8,Zeroquant} is widely recognized as one of the most effective techniques for addressing this challenge. By aggressively reducing the precision of LLM parameters from 16-bit floating-point (FP16) to low-bit integers such as 4-bit, both memory usage and computational complexity can be significantly decreased during inference. Quantization methods are broadly divided into Post-Training Quantization (PTQ)\citep{ptq} and Quantization-Aware Training (QAT)\citep{qat}. This work focuses on PTQ, due to the substantial computational cost of retraining LLMs required by QAT. For low-bit integer PTQ of LLM, the key challenge is a significant accuracy drop. According to previous research \citep{smoothquant, AWQ, rptq, quarot}, quantizing activations to low-bit integers is more challenging than quantizing weights due to the wider dynamic range and the presence of outliers in activations. As a result, these studies have proposed various methods to identify and mitigate activation outliers during the calibration phase of PTQ. 

For instance, data redistribution techniques \citep{smoothquant, qllm} have been widely adopted in LLMs to minimize quantization errors in activation data. However, these techniques still suffer from significant accuracy degradation when both weights and activations are quantized to 4-bit integers. Recently, a study utilizing rotation matrices \citep{quarot} successfully quantized both weights and activations to 4-bit integers while maintaining reasonable accuracy in LLMs. Specifically, Quarot \citep{quarot} utilizes orthogonal Hadamard-based rotations \citep{hadamard} to suppress outliers in activations and KV caches, effectively mitigating the impact of outliers in activations. However, Hadamard-based rotation does not always guarantee effective outlier suppression despite its computation overhead. Furthermore, since the efficiency of these rotations is highly dependent on the underlying data distribution, applying Quarot to models with multiple data distributions, such as in mixture-of-expert models, requires significant tuning efforts.


This paper introduces QRazor, a PTQ method to deliver reliable 4-bit LLMs. The QRazor employs two key insights:  
\begin{itemize}
    \item Utilizing 8-bit integers for weights and KV cache values, combined with 16-bit integers for activations, effectively captures data distributions, including outliers, and serves as the base precision scenario in this work.
    \item By capturing a few salient bits from the base precision scenario, we can preserve essential characteristics of both outliers and non-outliers at reduced bit precision.
\end{itemize}

Building on these insights, QRazor operates in two stages: quantization and compression. In the quantization stage, we first quantize weights, activations, and KV caches according to the base precision scenario described earlier. Our experiments demonstrate that in this PTQ setup, the accuracies of various LLMs remain nearly identical to those of their original counterparts. Next, the parameters are compressed to our target precision using our significant data razoring (SDR) technique. Our SDR technique captures a few salient bits and discards the other bits, implemented through bitwise operations, truncation, and round-to-nearest. Such a scheme does not manipulate data distributions and is easily adaptable to various models. 

This work considers the following two scenarios based on our QRazor scheme: W4A4 (4-bit weights, 4-bit activations, FP16 KV cache) and W4A4KV4 (4-bit weights, 4-bit activations, 4-bit KV cache). The accuracies of these configurations are compared against state-of-the-art (SOTA) 4-bit LLMs. For the LLaMA-1-7B and LLaMA-13B models, QRazor demonstrates superior performance, achieving more than a 10\% improvement in accuracy compared to QLLM \cite{qllm}. When compared to Quarot, QRazor achieves significantly better results when evaluated against the rounding-to-nearest (RTN) baseline, and nearly matching results for configurations involving GPTQ for the LLaMA-2-7B and LLaMA-13B models. For other model, such as Mistral-7B, QRazor achieves strong accuracy results. These findings demonstrate QRazor’s ability to deliver consistent and reliable performance across diverse LLM models.

Additionally, we develop an integer-based arithmetic unit specifically optimized for QRazor, enabling decompression-free computations and eliminating the hardware overhead associated with group-level dequantization. The multiplication of 4-bit compressed data is performed using a 4-bit integer multiplier, significantly improving computational efficiency. Hardware simulations show that our design, implemented for scenarios such as W4A4 or W4A4KV4, achieves 57.8\% power and 61.2\% area savings compared to arithmetic operations performed after decompression.

\section{Related Work}
\label{sec:related}
\textbf{LLM quantization.} 
Due to the wide quantization range required for activations, most previous PTQ schemes have focused on weight-only quantization, where weights are reduced to extremely low bits while activations remain in FP16 format \citep{gptq, spqr, owq, squeezellm, qulp, signround}. These schemes are optimized to minimize memory consumption, yet computations are still performed in high precision. To reduce the precision of both weights and activations, various studies have been conducted on mitigating activation outliers by balancing the data range \citep{outliersuppression, mofq, fptq, quantease, omniquant, normtweaking}. Furthermore, more fine-grained quantization strategies, such as channel-wise quantization \citep{outliertune} and group-wise quantization \citep{vsquant, Zeroquant}, have been explored to achieve reliable results in LLM services.

Recently, several works have succeeded in quantizing both weights and activations to 4-bit integers \citep{olive, quarot}. Atom \citep{atom} and QUIK \citep{quik} achieve this by quantizing most of the data to 4 bits while retaining a small portion of data in 8-bit form. QLLM \citep{qllm} introduces a channel reassembly technique, and QuaRot \citep{quarot} utilizes Hadamard-based rotations to suppress outlier issues and enable effective 4-bit inference.
Meanwhile, KV cache quantization is another crucial research topic, especially for LLMs \citep{flexgen, kvquant}, as the cache size becomes a major memory bottleneck when operating with large batches or generating long contexts. Notably, QuaRot \citep{quarot} demonstrates 4-bit quantization of all the data, including the KV cache.

\textbf{Outlier mitigation.} Recent studies have shown that smoothing activation outliers by transitioning magnitudes between activations and weights can effectively reduce the wide quantization range \citep{smoothquant}. Several studies have mitigated outliers using reorder-based quantization by clustering the activation channels \citep{rptq, atom}. These studies have reduced addressing complexity by fusing the reorder operation into the previous layer normalization operation. Other approaches, which apply transformations to the matrices to reduce outlier magnitudes, have also succeeded in reducing the entire weight and activation data to low-bit integers \citep{qllm, quarot}.

There are several encoding methods equipped with dedicated hardware support that effectively manage outliers to be processed with the same precision as non-outlier data \citep{o2a}. OliVe \citep{olive} proposes the outlier-victim pair mechanism, which provides extra representation space for outliers, thereby maintaining the same precision for all data. O2A \citep{o2a} encodes the entire data into low precision by leveraging additional flag bits. These flag bits, produced per group, indicate the magnitude of the outlier values. The uniformly compressed bits are then dequantized based on the flag bits for computation.


\section{Preliminaries}
\label {sec:preliminaries}
\textbf{Absolute Max Scaling.} 
The absolute max scaling \citep{absmax} ensures that input data fits within a target bit-width by utilizing the absolute maximum value of the data. The quantization process begins by identifying the absolute maximum value $\vert X_{\text{max}} \vert$ in the tensor, which is then used to calculate the scale factor. For example, in the case of 8-bit integer quantization, the formula to quantize a tensor $\displaystyle X$ under absolute max scaling and its dequantization formula are given by:

\vspace{-0.1cm}
\begin{center}
$\displaystyle X_q$ = round($\displaystyle\frac{127} {\vert X_{\text{max}} \vert}$ $\displaystyle \cdot X$), $\quad$ $\displaystyle \hat{X}$ = $\displaystyle\frac{\vert X_{\text{max}} \vert} {127}$$\displaystyle \cdot  X_q$
\end{center}
\vspace{-0.2cm}

where $\displaystyle X_q$ and $\displaystyle \hat{X}$ are the quantized and dequantized tensors, respectively. This scaling guarantees that all values in the tensor are normalized within the 8-bit range, maximizing the precision within the allowed bit-width. 
The absolute max scaling is applied to establish the base precision scenario outlined in Section \ref{sec:intro}, as it is known to be less sensitive to outliers compared to the min-max scaling method \citep{normalization,scaling}.

\begin{figure}[t]
\begin{center}
\centerline{\includegraphics[width=0.5\textwidth]{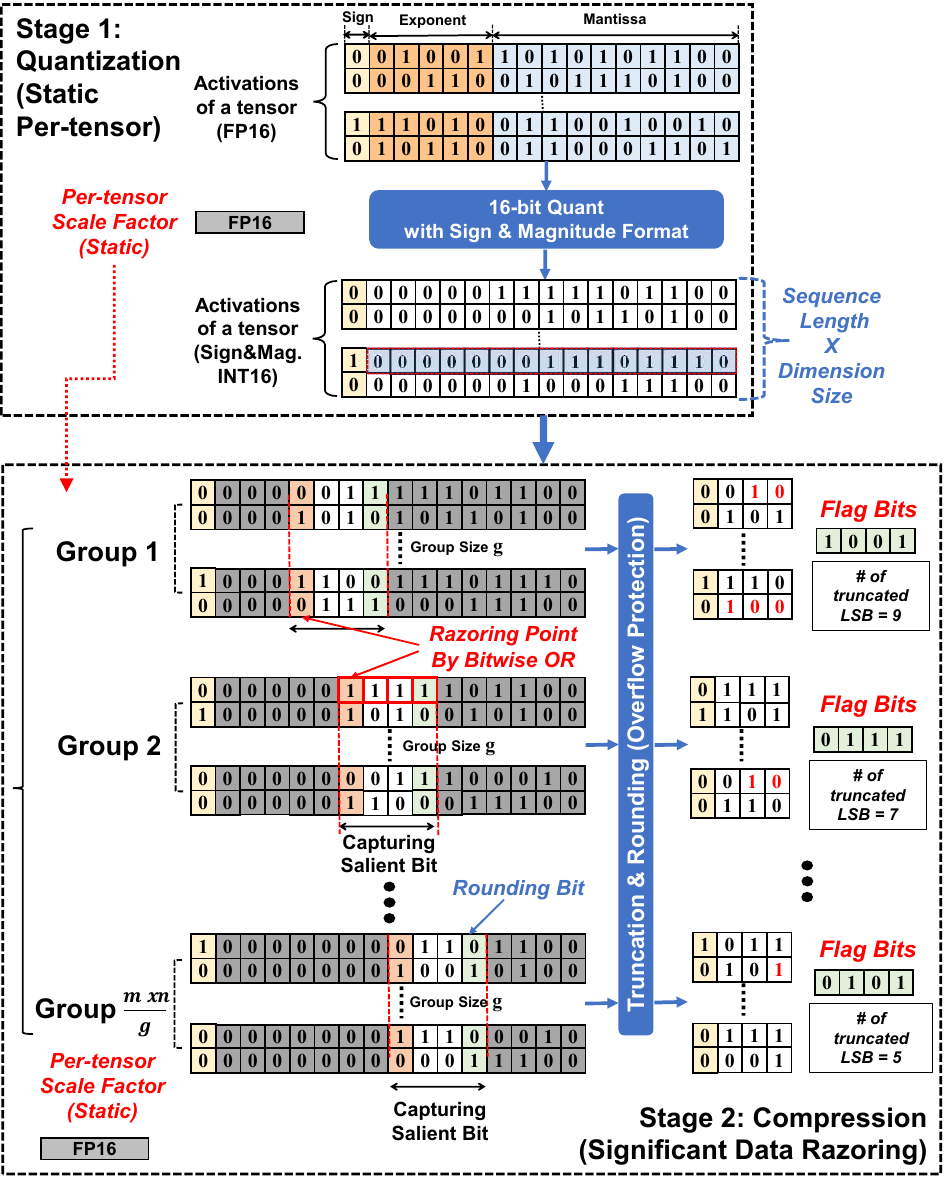}}
\vspace{-0.3cm}
\caption{The overall concept of our QRazor scheme.}
\label{fig:QRazor}
\end{center}
\vskip -0.2in
\end{figure}

\section{The QRazor Scheme}
\label {sec:QRazor}
Our QRazor scheme consists of two stages: quantization and compression. In the quantization stage, we first quantize FP16 parameters to high-bit integers to accurately represent LLM parameters without any loss of accuracy, which we refer to as the base precision scenario. It is important to note that, at this stage, only the static quantization method \citep{smoothquant} has been employed. Following this, the integers in the base precision format are compressed to lower-bit data using our SDR technique.

Figure \ref{fig:QRazor} illustrates the concept of our QRazor scheme, specifically detailing the process of quantizing activations to 4-bit. Most 4-bit PTQ schemes divide the tensor data into multiple groups. After grouping, the data in each group (initially in FP16 precision) are directly quantized to 4-bit using a single scale factor with FP16 precision. In contrast, our QRazor scheme quantizes the entire tensor data to 16-bit integers during the quantization stage, with a single scale factor shared across the entire tensor. Afterward, the quantized tensor is divided into multiple groups, and the 16-bit integer data in each group are compressed to 4-bit using our SDR technique.

It is important to emphasize that in the quantization stage of our QRazor scheme, activations and KV-caches utilize per-tensor scaling, respectively, while weights employ per-channel scaling. All parameters use static scaling at this stage. In contrast, Quarot, the SOTA method for quantizing all LLM parameters to 4-bit, similarly uses per-channel scaling for weights but adopts dynamic per-token scaling for activations and per-group scaling for KV caches, with a group size of 128, to enhance LLM accuracy in low-bit precision scenarios \citep{quarot}.

Unlike weights, which are quantized offline, activations and KV caches must be quantized online using static scaling parameters. In this context, the scaling granularity of activations and KV caches plays a critical role in determining the computational complexity of LLM inference. Our coarse granularity scaling approach—per-tensor for activations and KV caches—reduces the computational complexity compared to methods employing fine granularity scaling, such as per-token for activations and per-group for KV caches in Quarot \citep{smoothquant}. This coarse granularity effectively offsets the relative overhead introduced by compression, as discussed in the following subsections. Moreover, despite the inherent limitations of coarse granularity, our QRazor scheme achieves accuracies that are either superior or comparable to Quarot, as discussed in Section \ref{subsec:llm_result}. 

\subsection{Quantization Stage: W8A16 or W8A16KV8} \label {subsec:Stage1}

In the quantization stage, LLM parameters are converted from FP16 to two base precision scenarios, depending on whether KV cache compression is applied: W8A16 (8-bit weights, 16-bit activations) and W8A16KV8 (8-bit weights, 16-bit activations, 8-bit KV caches). These base precisions serve as primitive data types in the subsequent compression stage, where they are finally converted to 4-bit integer representation.

The base precision selected for per-tensor static quantization aims to preserve outliers within the tensors without employing additional techniques such as data redistribution or rotation. It is well known that activations typically exhibit a broader dynamic range than weights or KV caches \citep{qserve}, leading to more extreme outliers. Consequently, effective activation quantization requires a higher-precision format, such as 16-bit integers, to capture the full range of values while minimizing information loss.

Table \ref{w8a16table} presents the LLM accuracy results for three different precision scenarios, applying only per-tensor static quantization. The results strongly support the need for a wider data range in activations, as the scenario quantized with 16-bit activations achieves nearly identical accuracy to its FP16 counterpart across various LLaMA tasks, whereas W8A8 leads to significant accuracy degradation. These findings highlight the necessity of 16-bit integers for capturing the outlier-prone dynamic range of activations, while 8-bit integers are sufficient for weights and KV caches. Note that our razoring scheme, applied during the compression stage, involves a precise process of detecting salient data at the bit level, directly derived from the base integer data. Therefore, selecting the appropriate base precision during the quantization stage is crucial for preserving data accuracy.

\begin{table}[t]
\caption{Zero-shot accuracy of different base precision settings.}
\label{w8a16table}
\begin{center}
\scriptsize
\setlength{\tabcolsep}{6pt}
\begin{tabular}{l c c c c c c}
\toprule
\multicolumn{2}{c}{} & \multicolumn{5}{c}{\textbf{Zero-shot Accuracy} $\uparrow$} \\  
\cmidrule(lr){3-7} 
\multicolumn{1}{c}{\textbf{Model}} & \multicolumn{1}{c}{\textbf{\#Bits}} & 
\multicolumn{1}{c}{\textbf{PQ}} & \multicolumn{1}{c}{\textbf{AE}} & \multicolumn{1}{c}{\textbf{AC}} & \multicolumn{1}{c}{\textbf{HS}} & \multicolumn{1}{c}{\textbf{WG}}\\
\midrule

\multirow{4}{*}{\makecell{LLaMA-2\\-7B}} 
    & FP16    & 79.13     & 74.39 & 45.97 & 76.21 & 69.30 \\
    & W8A8       & 71.89 & 65.99 & 33.11 & 65.11 & 64.33 \\
    & W8A16      & 77.69 & 74.05 & 42.43 & 75.97 & 69.30 \\
    & W8A16KV8   & 78.07 & 74.20 & 42.41 & 76.26 & 69.30 \\
\midrule

\multirow{4}{*}{\makecell{LLaMA-2\\-13B}} 
    & FP16       & 80.23 & 77.82 & 48.76 & 79.39 & 72.30 \\
    & W8A8       & 68.99 & 40.35 & 27.09 & 57.26 & 50.99 \\
    & W8A16      & 78.94 & 76.47 & 45.14 & 78.92 & 71.27 \\
    & W8A16KV8   & 78.73 & 76.38 & 44.81 & 79.03 & 71.51 \\

\bottomrule
\end{tabular}
\end{center}
\end{table}

\subsection{Compression Stage: Significant Data Razoring} \label {subsec:Stage2}
The bottom part of Figure \ref{fig:QRazor} illustrates the proposed compression stage. Using sign-and-magnitude formatted integers obtained from the quantization stage, we apply our on-the-fly compression technique, termed significant data razoring (SDR). This method seamlessly integrates with the quantization stage and enables decompression-free computation, thereby minimizing the dequantization burden.

Our SDR technique identifies the “razoring point” for each group, which corresponds to the bit position of the leading one. The razoring point is determined by detecting the bit position of the leading one from the bitwise OR result of all data within the group (See Appendix \ref{appendix3} for details). Once the razoring point is identified, a defined number of adjacent bits starting from this point, known as the salient bits, are captured. The bit width of these salient bits matches the target precision. For instance, to achieve 4-bit activations, the width of the salient bits must be four, as illustrated in Figure \ref{fig:QRazor}.

\rule{\linewidth}{0.4mm}
\textbf{Algorithm 1} Pseudo Code of QRazor Compression 
\hrule
\scriptsize
$\displaystyle func$ QRazor$\displaystyle (X_{i,j}, w, t, g)$ \\ 
\begin{tabular}{ll}
{\bf Input}: \\
\hspace{2mm}$\displaystyle X_{i,j} \in \displaystyle \mathbb{R}$ --- Weight, Activation, KV cache matrix, \\
\hspace{2mm}$\displaystyle b_w$          --- Quantization bit width \\
\hspace{2mm}$\displaystyle b_t$          --- Total bits to remove for compression\\
\hspace{2mm}$\displaystyle g$          --- Element group size \\

{\bf Initialize}: \\
\hspace{2mm}$\displaystyle b_k=b_w-b_t$ \hspace{10mm} // compressed bit width\\

{\bf Algorithm Steps}: \\
\hspace{2mm}$\displaystyle E_{i,j} = $ Quantize($\displaystyle X_{i,j}, b_w$) \\
\hspace{2mm}For each element  $\displaystyle e_{i,j}$ in $\displaystyle E_{i,j}$ \\
\hspace{2mm}$\displaystyle s_{i,j} \leftarrow$ sign bit of $\displaystyle e_{i,j}$ \hspace{26mm} \\
\hspace{2mm}{\bf for} $\displaystyle i = 1,2,...,m$ {\bf do} \\
\hspace{4mm}{\bf for} $\displaystyle j = 1,2,...,n$ {\bf do} \\
\hspace{6mm}{\bf if} $\displaystyle e_{i,j} < 0$ {\bf do} \\
\hspace{8mm}$\displaystyle e_{\text{si}_{i,j}} = $ remove\_sign\_bit($\displaystyle e_{i,j}$) \\
\hspace{8mm}$\displaystyle e_{\text{ci}_{i,j}} = $ concat($\displaystyle s_{i,j}$, 2's\_complement($\displaystyle e_{\text{si}_{i,j}}$)) \\
\hspace{6mm}{\bf else} {\bf do} \\
\hspace{8mm}$\displaystyle e_{\text{ci}_{i,j}} =  e_{i,j}$ \\
\hspace{2mm}For each element  $\displaystyle e_{\text{ci}_{i,j}}$ in $\displaystyle E_{\text{ci}_{i,j}}$ \\
\hspace{2mm}$\displaystyle G_p = $ group\_by\_size\_$g$($\displaystyle e_{\text{ci}_{i,j}},g$) \\
\hspace{2mm}{\bf for} $\displaystyle p = 1,2,...,\displaystyle\frac{m \cdot n} {g}$ {\bf do} \hspace{15mm} // ($\displaystyle\frac{m \cdot n} {g} \in \displaystyle Z^+$) \\
\hspace{4mm}$\displaystyle b_m =$ detect\_redundant\_MSBs($\displaystyle G_p, b_t, b_w$ \\
\hspace{4mm}$\displaystyle b_l=b_t-b_m$ \\
\hspace{4mm}$\displaystyle F_{flagbit} = b_l$ \\
\hspace{2mm}$\displaystyle G_q = $ truncate\_MSBs\_and\_LSBs($\displaystyle e_{\text{ci}_{i,j}},b_m,b_l$) \\ 
\hspace{4mm}{\bf if} $\displaystyle G_q = 2^k - 1$ {\bf do} \\
\hspace{4mm}$\displaystyle G_{r} \leftarrow G_{q}^{\text{floor}}$ \hspace{10mm} // truncate LSBs without carry\\
\hspace{4mm}{\bf else} {\bf do} \\
\hspace{6mm}{\bf if} $\displaystyle \text{MSB of truncated LSBs}$ $\displaystyle = 0$ {\bf do} \\
\hspace{8mm}$\displaystyle G_{r} \leftarrow G_{q}^{\text{floor}}$ \\
\hspace{6mm}{\bf else} {\bf do} \\
\hspace{8mm}$\displaystyle G_{r} \leftarrow G_{q}^{\text{ceil}}$ \hspace{7mm} // $\displaystyle r \in \{0, 1, 2, \dots, \frac{m \cdot n}{g}\}$ \\
\hspace{2mm} $\displaystyle C= \{G_1, G_2, \dots, G_{\frac{m \cdot n}{g}}\}$ \\
\hspace{1mm} {\bf return} $\displaystyle C, F_{flagbit}$ \\
\end{tabular} 
\hrule
\normalsize

\vspace{0.3cm}
After capturing salient bits, we retain only the sign bit and the selected salient bits, truncating all the other bits. Then, by rounding the last bit of the remaining bits, we obtain the final compressed values. In such a scheme, during the rounding process of the least significant bits (LSBs), there is a potential risk of overflow. For example, rounding up ``$\displaystyle 01111_2$," where the most significant bit (MSB) represents the sign, could alter the sign bit. To prevent this, we avoid rounding the LSBs of elements where all salient bits are `1'. Instead, we apply flooring to the LSBs of these elements while continuing to round the LSBs of other elements within the group. The detailed procedure of our QRazor scheme, including the aforementioned exception handling, is outlined in Algorithm 1.

Ultimately, our SDR technique eliminates the `0's in higher bit positions above the razoring point, which can be reconstructed using flag bits that indicate the razoring point for each group — the flag bits represent the number of truncated LSBs in the group and thus inform the razoring point when the base precision is known. We validate the efficacy of our SDR technique across various scenarios, demonstrating that it effectively compresses all data into a 4-bit format while delivering reliable accuracies across multiple LLM tasks, as further discussed in Section \ref{subsec:llm_result}.

The size of the compression group is determined within the range of 16 to 128. The reason we can accommodate relatively small group sizes during compression is that our SDR technique does not involve any computational quantization but simply removes zeros from the MSB portion of the integer-based data, along with a straightforward rounding process. As a result, unlike other group-based quantization approaches, there is no need for a scaling process per group using higher precision. This enables direct computation with low-bit operands, supported by our dedicated arithmetic design. The design facilitates both memory- and compute-efficient low-bit matrix multiplication, with hardware details provided in Section \ref{subsec:arith}. It is important to note that the FP16 scaling process in QRazor is applied only during the quantization stage, with granularities of per-tensor for activations and KV cache while per-channel for weights.

\begin{figure}[t]
\begin{center}
    \includegraphics[width=1.00\columnwidth]{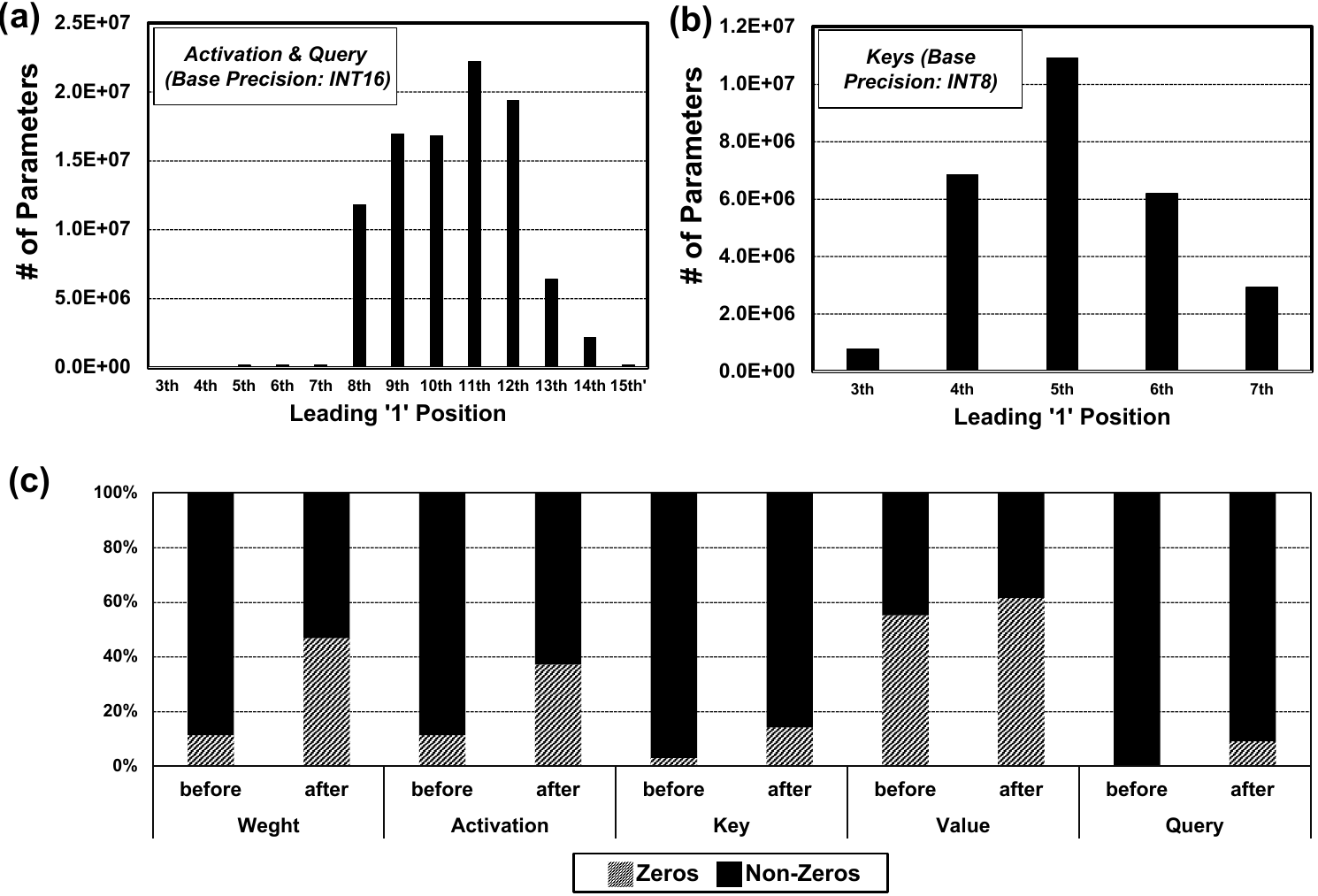}
\end{center}
\vspace{-0.3cm}
\caption{Leading '1' position before the compression for (a) activation and query, and (b) key. (c) portion of zeroed elements between ``Before 4-bit compression" and ``After 4-bit compression"}
\label{Leading1}
\end{figure}

Due to the aggressive 4-bit compression, one might be concerned that the number of zeroed elements significantly increases, degrading the reliability of our QRazor. Figure \ref{Leading1}(a), (b) demonstrates that the first concern is effectively addressed in our QRazor. It illustrates the portion of leading `1' positions before 4-bit compression (immediately after converting to the sign-and-magnitude format). The leading `1' positions for activations are predominantly located between the 8th and 12th bit orders. For instance, if the leading `1' position in a group is the 13th bit, parameters with an MSB below the 9th bit will be rounded and zeroed after compression, as only 4 bits are retained. The empirical results show that, the number of groups where the leading `1' position exceeds the 12th-bit order is minimal (only 9\%), indicating that outliers are infrequent and mitigating concerns about significant parameters being truncated after 4-bit compression.

\vspace{-0.2cm}

In Figure \ref{Leading1}(c), we also analyzed the proportion of zeroed elements before and after 4-bit compression. For Query, Key, and Value, the increase in zeroed elements is not so considerable. However, for activations and weights, there is a notable increment. This is expected, as significant activations and weights often have small absolute values close to zero. Such truncation of small values does not lead to substantial errors. Combined with the low occurrence of outliers, as mentioned above, this explains why QRazor consistently delivers reliable performance across various LLMs, as discussed in Section \ref{subsec:llm_result}.

It is important to note that in our compression, the truncation level dynamically varies for each group during runtime. Due to this characteristic, one might draw a comparison between our compression and dynamic max-scaled quantization (DMQ). However, our compression technique fundamentally differs from DMQ in the following aspects:

\textbf{Eliminating Absolute Max Computation:} Instead of determining the absolute maximum value within a group, QRazor detects only the leading '1'. While the absolute maximum inherently contains the leading '1', multiple parameters may share the same leading '1' position. In DMQ, identifying the absolute maximum value which is presented in floating point data format manner, is essential for computing the scaling factor. In contrast, our compression bypasses this step entirely, significantly reducing computational complexity.

\textbf{Lightweight Compression and Decompression:} Per-group DMQ typically involves group-level quantization and dequantization operations, which rely on arithmetic computations such as multiplication and, in some cases, division. In contrast, our compression and decompression rely on bit-level truncation and shifting, which are inherently simpler and far less resource-intensive.

\begin{figure*}[t]
\begin{center}
\centerline{\includegraphics[height=0.257\textwidth]{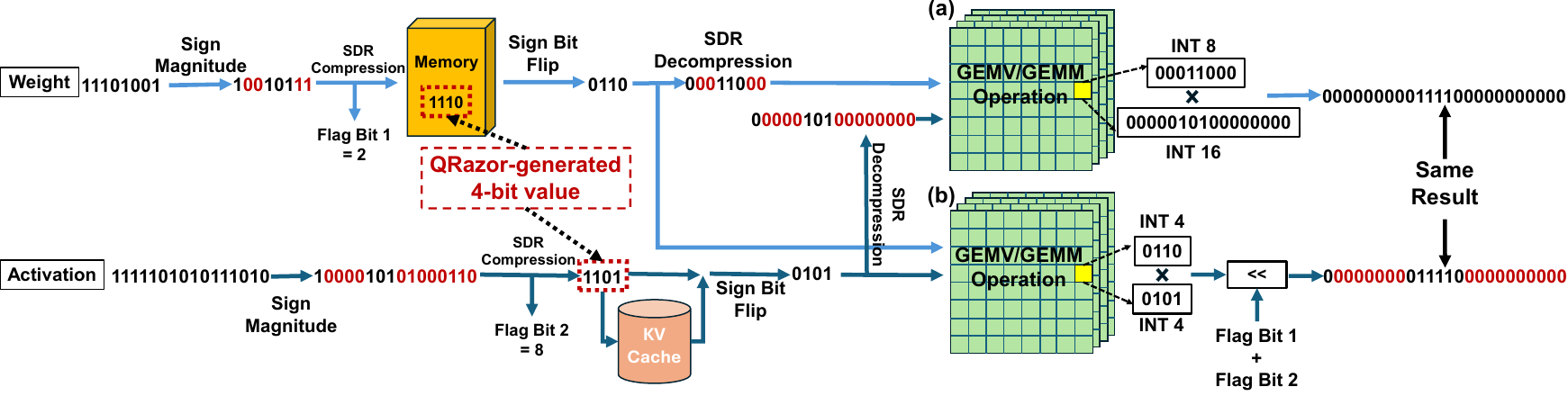}}
\vspace{-0.3cm}
\caption{The concept of  a commonly perceived decompression method(a) compared with the proposed decompression-free arithmetic(b).}
\label{fig:decomp}
\end{center}
\vskip -0.2in
\end{figure*}

To conclude, while DMQ and our compression share the goal of dynamically reducing precision, their underlying methodologies are fundamentally different. As outlined above, QRazor’s compression incurs substantially lower computational overhead than DMQ. Moreover, QRazor’s simpler operations require fewer hardware resources, enabling more efficient implementation in dedicated arithmetic units, as further discussed in the following section.

\subsection{Decompression-Free Arithmetic Operation} \label {subsec:arith}
We introduce an arithmetic unit designed to execute multiply-and-accumulate (MAC) operations required for matrix multiplication while preserving the compression of our QRazor, a method we refer to as decompression-free arithmetic. Typically, data compressed by QRazor would need to be decompressed to their base precisions before performing MAC operations. This additional decompression process can introduce significant area overhead and degrade throughput, ultimately reducing performance when deployed on hardware devices. To address this challenge, we propose a decompression-free integer-based arithmetic unit, which enhances throughput by directly computing with the compressed low-precision data. This is achieved and evaluated by implementing the dedicated processing unit using a 4$\times$4 multiplier and a single 16-bit barrel shifter.

The top part of Figure~\ref{fig:decomp} illustrates the concept of conventional arithmetic operations for QRazor, which involve decompression of operands before computation. In this approach, weight and activation values are decompressed separately to their base precision using logical shift operations, followed by multiplication with a high-precision of 16$\times$8 multiplier. In contrast, the proposed decompression-free arithmetic operation is shown in the bottom part of Figure~\ref{fig:decomp}. Here, the operands are directly multiplied using the 4$\times$4 multiplier. The multiplied results are then manipulated by the flag bits using a single 16-bit barrel shifter. This approach eliminates the need for separate decompression steps, thereby enhancing computational efficiency and throughput. The hardware efficiency is further analyzed in Section \ref{subsec:hardware}.

\section{Experiments} \label {sec:experiments}
\subsection{Setup}
We evaluate our proposed QRazor method on the LLaMA2\citep{llama2} and LLaMA3\citep{llama3} models. Static quantization utilizing the absolute max scaling is implemented for all weights, activations, and KV cache values, with granularity configured as per-channel for weights while per-tensor for activations and KV cache values. The SDR technique is applied for 4-bit compression using various group sizes. Weight compression is performed offline, while activations and KV caches are compressed online. Calibration and validation of zero-shot accuracy and perplexity were performed by randomly selecting \texttt{128} samples from \texttt{Wikitext2} for all tasks, with perplexity evaluated on the same task while accuracy on \texttt{PIQA}, \texttt{ARC}, \texttt{Hellaswag}, and \texttt{Winogrande} using \texttt{lm-evaluation-harness} \citep{eval-harness}. Additionally, an ablation study was conducted to assess the impact of group size on quantization accuracy, focusing on the compressed bits for weights and activations. A sequence length of \texttt{2048} and a batch size of \texttt{1} were used for all tasks during evaluation, and the models provided by \texttt{meta-llama} and \texttt{mistralai-mistral} were employed for the tests.

\subsection{LLM Results} \label {subsec:llm_result}
\textbf{LLaMA-2 and LLaMA-3:} We conduct comprehensive experiments to evaluate the performance of both LLaMA-2 and LLaMA-3 across various model scales. Table \ref{maintable} presents the overall accuracy results, comparing our QRazor scheme with previous SOTA quantization methods that provide W4A4 precision. Notably, both our approach and QuaRot \citep{quarot} additionally offer 4-bit KV cache quantization. For our QRazor scheme, group sizes of 16 and 32 are evaluated, where groups share the same flag bits. These group sizes yield the same effective bits per data as conventional group-wise quantization schemes with 128 data per group. As previously mentioned, our scheme enables direct low-precision processing, ensuring that smaller group sizes hardly degrade hardware performance, as no dequantization operations are required across groups.

\begin{table*}[t]
\vspace{-0.3cm}
\caption{Zero-shot accuracy of LLaMA families and Mistral-7B on five common sense tasks.}
\label{maintable}
\renewcommand{\arraystretch}{0.6}
\begin{center}
\scriptsize
\setlength{\tabcolsep}{6pt}
\begin{tabular}{ccccccccccc}
\toprule
\multicolumn{4}{c}{} & 
\multicolumn{1}{c}{\textbf{Perplexity} $\downarrow$}&
\multicolumn{6}{c}{\textbf{Zero-shot Accuracy} $\uparrow$} \\
\cmidrule(lr){5-5}
\cmidrule(lr){6-11} 
\multicolumn{1}{c}{Model} & \multicolumn{1}{c}{\#Bits} & \multicolumn{1}{c}{Eff. Bits} & \multicolumn{1}{c}{Method} & 
\multicolumn{1}{c}{Wikitext2} &
\multicolumn{1}{c}{PIQA} & \multicolumn{1}{c}{ARC-e} & \multicolumn{1}{c}{ARC-c}  & \multicolumn{1}{c}{HellaSwag} & \multicolumn{1}{c}{Winogrande} & \multicolumn{1}{c}{Avg} \\

\cmidrule(lr){1-11}

\multirow{15}{*}{LLaMA-2-7B} & { FP16} & { 16} & { -} & { 5.47} & { 79.13} & { 74.39} & { 45.97} & { 76.21} & { 69.30} & { 69.00} \\

\cmidrule(lr){2-11}

{ } & { W4A4 } & { 4.03/4} & { OS+} & { -}& { 63.11} & { 39.10} & { 28.84} & { 47.31} & { 51.30} & { 45.93} \\

{ } & { W4A4} & { 4} & { OmniQuant} & { 14.61} & { 65.94} & {43.94} & { 30.80} & { 53.53} & {55.09} & { 49.86} \\

{ } & { W4A4} & { 4} & { QLLM} & { 11.75} & { 67.68} & { 44.40} & { 30.89} &{ 58.45} & { 56.59} & { 51.60}  \\

\cmidrule(lr){2-11}

{ } & { W4A4KV4} & \multirow{2}{*}{ 4/4/4.125} & { QuaRot(RTN)} & { 8.37} & { 72.09} & { 58.88} & { 35.24} & { 65.40} & { 60.69} & { 58.26} \\

{ } & { W4A4KV4} & {} & { QuaRot(GPTQ)} & { 6.10} & { 76.77} & { 69.87} & { 40.87} & { 72.16} & { 63.77} & { 65.64} \\

\cmidrule(lr){2-11}

{ } & { W4A4 $\displaystyle g16$} & { 4.25} & { } & \cellcolor{blue!10}{ 6.45} & \cellcolor{blue!10}{ 75.84} & \cellcolor{blue!10}{ 72.63} & \cellcolor{blue!10}{ 42.47} & \cellcolor{blue!10}{ 72.96} & \cellcolor{blue!10}{ 65.67} & \cellcolor{blue!10}{ 65.91} \\

{ } & { W4A4 $\displaystyle g32$} & { 4.125} & \multirow{2}{*}{ QRazor} & \cellcolor{blue!10}{ 7.03} & \cellcolor{blue!10}{ 73.67} & \cellcolor{blue!10}{ 70.70} & \cellcolor{blue!10}{ 39.46} & \cellcolor{blue!10}{ 71.46} & \cellcolor{blue!10}{ 64.09} & \cellcolor{blue!10}{ 63.88} \\

{ } & {W4A4KV4 $\displaystyle g16$} & { 4.25} & { } & \cellcolor{blue!10}{ 7.62} & \cellcolor{blue!10}{ 73.39} & \cellcolor{blue!10}{ 70.88} & \cellcolor{blue!10}{ 39.80} & \cellcolor{blue!10}{ 70.15} & \cellcolor{blue!10}{ 64.01} & \cellcolor{blue!10}{ 63.65} \\

{ } & { W4A4KV4 $\displaystyle g32$} & { 4.125} & { } & \cellcolor{blue!10}{ 8.14} & \cellcolor{blue!10}{ 73.23} & \cellcolor{blue!10}{ 67.54} & \cellcolor{blue!10}{ 37.46}  & \cellcolor{blue!10}{ 67.16} & \cellcolor{blue!10}{ 60.46} & \cellcolor{blue!10}{ 61.17} \\

\cmidrule(lr){1-11}

\multirow{15}{*}{ LLaMA-2-13B} & { FP16} & { 16} & {-} & { 4.88} & { 80.23} & { 77.82} & { 48.76} & { 79.39} & { 72.30} & { 71.70} \\

\cmidrule(lr){2-11}

{ } & { W4A4} & { 4.03/4} & { OS+} & { -} & { 64.47} & { 41.46} & { 32.17} & { 59.30} & { 51.38} & { 49.76} \\

{ } & { W4A4} & { 4} & { OmniQuant} & { 12.28} & { 69.80} & { 47.22} & { 33.79} & { 59.34} & { 55.49} & { 53.13} \\

{ } & { W4A4} & { 4} &  { QLLM} & { 9.09} & { 70.46} & { 48.48} & { 34.39} & { 62.80} & { 55.41} & { 54.31}  \\

\cmidrule(lr){2-11} 


{ } & { W4A4KV4} & \multirow{2}{*}{ 4/4/4.125} & { QuaRot(RTN)} & { 6.09} & { 77.37} & { 70.83} & { 43.69} & { 73.11} & { 67.32} & { 67.16} \\

{ } & { W4A4KV4} & {} & { QuaRot(GPTQ)} & { 5.40} & { 78.89} & { 72.98} & { 46.59}  & { 76.37} & { 70.24} & { 69.79} \\

\cmidrule(lr){2-11} 

{ } & { W4A4 $\displaystyle g16$} & { 4.25} & { } & \cellcolor{blue!10}{ 5.37} & \cellcolor{blue!10}{ 77.48} & \cellcolor{blue!10}{ 75.09} & \cellcolor{blue!10}{ 43.47} & \cellcolor{blue!10}{ 76.83} & \cellcolor{blue!10}{ 70.09} & \cellcolor{blue!10}{ 68.59}\\

{ } & { W4A4 $\displaystyle g32$} & { 4.125} & \multirow{2}{*}{ QRazor} & \cellcolor{blue!10}{ 5.54} & \cellcolor{blue!10}{ 76.82} & \cellcolor{blue!10}{ 73.86} & \cellcolor{blue!10}{ 41.80} & \cellcolor{blue!10}{ 76.30} & \cellcolor{blue!10}{ 68.59} & \cellcolor{blue!10}{ 67.47} \\

{ } & { W4A4KV4 $\displaystyle g16$} & { 4.25} &  { } & \cellcolor{blue!10}{ 5.49} & \cellcolor{blue!10}{ 77.80} & \cellcolor{blue!10}{ 75.44} & \cellcolor{blue!10}{ 44.15} & \cellcolor{blue!10}{ 76.06} & \cellcolor{blue!10}{ 67.72} & \cellcolor{blue!10}{ 68.23}  \\

{ } & { W4A4KV4 $\displaystyle g32$} & { 4.125} & { } & \cellcolor{blue!10}{ 5.79} & \cellcolor{blue!10}{ 77.09} & \cellcolor{blue!10}{ 74.91} & \cellcolor{blue!10}{ 44.28} & \cellcolor{blue!10}{ 75.06} & \cellcolor{blue!10}{ 66.77} & \cellcolor{blue!10}{ 67.62} \\

\cmidrule(lr){1-11}

\multirow{10}{*}{ LLaMA-3-8B} & { FP16} & { 16} & {-} & { 6.14} & { 80.65} & { 80.18} & { 50.84} & { 79.18s} & { 73.01} & { 72.77} \\

\cmidrule(lr){2-11}






{ } & { W4A4KV4} & \multirow{2}{*}{ 4/4/4.125} & { QuaRot(RTN)} & { -} & { -} & { -} & { -} & { -} & { -} & { -} \\

{ } & { W4A4KV4} & {} & { QuaRot(GPTQ)} & { 8.16} & { 75.14} & { 68.01} & { 43.34}  & { 72.94} & { 72.77} & { 65.18} \\

\cmidrule(lr){2-11} 

{ } & { W4A4 $\displaystyle g16$} & { 4.25} & { } & \cellcolor{blue!10}{ 6.49} & \cellcolor{blue!10}{ 78.42} & \cellcolor{blue!10}{ 79.65} & \cellcolor{blue!10}{ 46.82} & \cellcolor{blue!10}{ 77.28} & \cellcolor{blue!10}{ 71.90} & \cellcolor{blue!10}{ 70.81}\\

{ } & { W4A4 $\displaystyle g32$} & { 4.125} & \multirow{2}{*}{ QRazor} & \cellcolor{blue!10}{ 7.05} & \cellcolor{blue!10}{ 78.24} & \cellcolor{blue!10}{ 77.89} & \cellcolor{blue!10}{ 45.82} & \cellcolor{blue!10}{ 76.93} & \cellcolor{blue!10}{ 71.40} & \cellcolor{blue!10}{ 70.06} \\

{ } & { W4A4KV4 $\displaystyle g16$} & { 4.25} &  { } & \cellcolor{blue!10}{ 7.57} & \cellcolor{blue!10}{ 77.42} & \cellcolor{blue!10}{ 77.19} & \cellcolor{blue!10}{ 44.15} & \cellcolor{blue!10}{ 75.81} & \cellcolor{blue!10}{ 69.30} & \cellcolor{blue!10}{ 68.77}  \\

{ } & { W4A4KV4 $\displaystyle g32$} & { 4.125} & { } & \cellcolor{blue!10}{ 8.26} & \cellcolor{blue!10}{ 76.77} & \cellcolor{blue!10}{ 75.51} & \cellcolor{blue!10}{ 42.81} & \cellcolor{blue!10}{ 74.05} & \cellcolor{blue!10}{ 68.75} & \cellcolor{blue!10}{ 67.58} \\

\cmidrule(lr){1-11}

\multirow{7}{*}{ Mistral-7B} & { FP16} & { 16} & {-} & { 5.42} & { 80.74} & { 81.58} & { 50.16} & { 61.25} & { 74.27} & { 69.60} \\

\cmidrule(lr){2-11}






{ } & { W4A4 $\displaystyle g16$} & { 4.25} & { } & \cellcolor{blue!10}{ 6.06} & \cellcolor{blue!10}{ 77.75} & \cellcolor{blue!10}{ 77.98} & \cellcolor{blue!10}{ 45.48} & \cellcolor{blue!10}{ 57.70} & \cellcolor{blue!10}{ 69.77} & \cellcolor{blue!10}{ 65.74}\\

{ } & { W4A4 $\displaystyle g32$} & { 4.125} & \multirow{2}{*}{ QRazor} & \cellcolor{blue!10}{ 6.24} & \cellcolor{blue!10}{ 77.80} & \cellcolor{blue!10}{ 75.39} & \cellcolor{blue!10}{ 43.47} & \cellcolor{blue!10}{ 56.20} & \cellcolor{blue!10}{ 68.90} & \cellcolor{blue!10}{ 64.35} \\

{ } & { W4A4KV4 $\displaystyle g16$} & { 4.25} &  { } & \cellcolor{blue!10}{ 6.18} & \cellcolor{blue!10}{ 77.78} & \cellcolor{blue!10}{ 77.34} & \cellcolor{blue!10}{ 44.48} & \cellcolor{blue!10}{ 58.09} & \cellcolor{blue!10}{ 68.56} & \cellcolor{blue!10}{ 65.25}  \\

{ } & { W4A4KV4 $\displaystyle g32$} & { 4.125} & { } & \cellcolor{blue!10}{ 6.93} & \cellcolor{blue!10}{ 77.53} & \cellcolor{blue!10}{ 76.09} & \cellcolor{blue!10}{ 42.24} & \cellcolor{blue!10}{ 56.38} & \cellcolor{blue!10}{ 66.38} & \cellcolor{blue!10}{ 63.72} \\

\bottomrule
\end{tabular}
\end{center}
\vspace{-0.1cm}
\scriptsize{
* Note: In the case of quantization granularity of channel and tensor, scale factor overhead is ignored.
}
\end{table*}

While QLLM \citep{qllm} preserves the acknowledgeable accuracies in LLaMA-2 models compared to previous methods, it still suffers from significant accuracy degradation. Additionally, QLLM does not address the quantization of KV caches into a low-precision format, which represents a significant memory bottleneck during long-context LLM inference. In contrast, our QRazor compresses KV caches to 4-bit while achieving higher accuracy than QLLM in LLaMA-2 models. For LLaMA-2 models, QuaRot is also compared, showing reasonable results with a marginal accuracy drop in most tasks using W4A4KV4 precision. When comparing the methods equipped with the rounding-to-nearest (RTN), commonly applied for both weight and activation compression, QRazor outperforms QuaRot, exhibiting an average accuracy drop of $<$8\% in LLaMA-2 models. QuaRot suggests that applying GPTQ \citep{gptq} to the weights could further reduce the accuracy drop. In this case, results with similar or slightly better accuracy are observed compared to our QRazor with RTN applied. However, for the LLaMA-3-8B model, QRazor provides better performances than QuaRot regardless of their weight quantization approaches. Since our scheme is solely influenced by the internal data distribution and is independent of other external criteria, various optimization techniques such as GPTQ can also be applied to our 4-bit model for further enhancements. We leave these potential improvements for future work.

\textbf{Mistral-7B:} In addition to the LLaMA models, we conducted an accuracy analysis on Mistral-7B achieving strong results across these models, as shown in Table~\ref{maintable}. These findings highlight QRazor’s capability to deliver reliable performance across various models and configurations.

\subsection{Ablation Studies} \label {subsec:ablation}
\textbf{W4A8 evaluation.} Although QRazor outperforms various 4-bit quantization approaches in task performance, some accuracy degradation is inevitable due to the aggressive low-precision representation. Therefore, we evalutate W4A8 version of QRazor by increasing the number of captured salient bits to 8 per activation.


\begin{table}[t]
\vspace{-0.4cm}
\caption{Zero-shot accuracies of W4A8 LLaMA2 models.}
\vspace{-0.3cm}
\label{w4a8table1}
\renewcommand{\arraystretch}{0.9}
\begin{center}
\tiny
\setlength{\tabcolsep}{2.5pt}
\resizebox{\textwidth}{!}{ }
\begin{tabular}{cccccccccc}
\toprule
\multicolumn{3}{c}{} & \multicolumn{6}{c}{\textbf{Zero-shot Accuracy} $\uparrow$} \\  
\cmidrule(lr){4-8} 
\multicolumn{1}{c}{Model} & \multicolumn{1}{c}{\#Bits} & \multicolumn{1}{c}{Method} & 
\multicolumn{1}{c}{PQ} & \multicolumn{1}{c}{AE} & \multicolumn{1}{c}{AC} & \multicolumn{1}{c}{HS} & \multicolumn{1}{c}{WG} & \multicolumn{1}{c}{Avg} \\

\midrule

\multirow{10}{*}{LLaMA-2-7B} & { FP16} & { -} & { 79.13} & { 74.39} & { 45.97} & { 76.21} & { 69.30} & { 69.00} \\

\cmidrule(lr){2-9}

{  } & { W4A8} & { QLLM} & { 76.11} & { 51.73} & { 39.33} & { 71.27} & { 65.59} & { 60.81}  \\

\cmidrule(lr){2-9} 

{  } & { W4A8KV4} & { QServe} & { 77.64} & { 72.81} & { 43.60} & { 74.00} & { 68.03} & { 67.22}  \\

{  } & { W4A8KV4 $\displaystyle g128$} & { QServe} & { 78.07} & { 73.32} & { 44.80} & { 74.98} & { 68.59} & { 67.95}  \\

\cmidrule(lr){2-9} 

{ } & { W4A8  $\displaystyle g16$ } & { } &{ 77.48} & { 72.39} & { 44.15} & { 74.87} & { 68.35} & { 67.45}  \\

{ } & { W4A8  $\displaystyle g32$} & \multirow{2}{*}{ QRazor} & { 77.09} & { 72.91} & { 44.15} & { 74.50} & { 68.98} & { 67.53} \\

{ } & { W4A8KV4  $\displaystyle g16$} & { } & { 76.55} & { 73.86} & { 43.14} & { 73.81} & { 68.59} & { 67.19} \\

{ } & { W4A8KV4  $\displaystyle g32$} & { } & { 76.66} & { 71.75} & { 43.82}  & { 72.97} & { 68.11} & { 66.66} \\

\midrule

\multirow{10}{*}{ LLaMA-2-13B} & { FP16} & {-} & { 80.23} & { 77.82} & { 48.76} & { 79.39} & { 72.30} & { 71.70} \\

\cmidrule(lr){2-9}

{ } & { W4A8} & { QLLM} & { 78.67} & { 57.11} & { 41.89} & { 75.33} & { 68.75} & { 64.35} \\

\cmidrule(lr){2-9} 

{  } & { W4A8KV4} & { QServe} & { 79.71} & { 75.97} & { 48.38} & { 77.80} & { 70.96} & { 70.56}  \\

{  } & { W4A8KV4 $\displaystyle g128$} & { QServe} & { 79.43} & { 77.06} & { 48.81} & { 78.35} & { 65.59} & { 70.83}  \\

\cmidrule(lr){2-9}

{ } & { W4A8  $\displaystyle g16$ } & { } & { 79.33} & { 76.42} & { 45.14} & { 78.24} & { 71.74} & { 70.17} \\

{ } & { W4A8  $\displaystyle g32$} & \multirow{2}{*}{ QRazor} & { 79.00} & { 76.07} & { 46.48} & { 78.07} & { 69.93} & { 69.91} \\

{ } & { W4A8KV4  $\displaystyle g16$ } & { } & { 78.51} & { 76.67} & { 45.15} & { 77.92} & { 71.19} & { 69.88} \\

{ } & { W4A8KV4  $\displaystyle g32$ } & { } & { 78.24} & { 76.84} & { 47.83} & { 77.54} & { 69.93} & { 70.07} \\

\bottomrule

\vspace{-0.5cm}
\end{tabular}
\end{center}
\end{table}

Table \ref{w4a8table1} presents the accuracy results under the W4A8 precision setting. The W4A8 configuration demonstrates significantly higher accuracies across various tasks than W4A4. At W4A8 precision, our QRazor consistently outperforms QLLM in most cases. Additionally, we further quantize KV caches to 4 bits, denoted as W4A8KV4, resulting in minimal accuracy loss. When comparing W4A8 to W4A4 in Table \ref{maintable}, QRazor demonstrates less degradation in performance with reduced activation precision compared to QLLM. Notably, our SDR technique captures salient bits at the bit level, enabling finer data preservation than other methods, which proves especially advantageous in more aggressive compression settings.

Compared to another SOTA method, QServe \cite{qserve}, QRazor achieves nearly comparable accuracies. However, QServe employs complex rotation operations for quantizing activations, similar to Quarot, which introduces significant computational overhead. This highlights that QRazor is a more efficient technique than QServe, at the W4A8KV4 configuration.

\begin{table}[t]
\caption{Zero-shot accuracy and effective bit-width comparison between different group sizes of W4A4KV4 configuration.}
\vspace{-0.3cm}
\label{grouptable}
\begin{center}
\setlength{\tabcolsep}{3pt}
\scriptsize
\begin{tabular}{cccccccc}
\toprule  
\multicolumn{1}{c}{Group ($\displaystyle g$) Size} &\multirow{3}{*}{\scriptsize Model}& \multirow{3}{*}{\scriptsize Baseline} & \multicolumn{1}{c}{8} & \multicolumn{1}{c}{16} & 
\multicolumn{1}{c}{32} & \multicolumn{1}{c}{64} & \multicolumn{1}{c}{128} \\

\cmidrule(lr){4-4} \cmidrule(lr){5-5} \cmidrule(lr){6-6} \cmidrule(lr){7-7} \cmidrule(lr){8-8}

{\scriptsize Effective Bits} & {\scriptsize } & {} & {\scriptsize 4.5} & {\scriptsize 4.25} & {\scriptsize 4.125} & {\scriptsize 4.06} & {\scriptsize 4.03} \\

\midrule 

\multirow{3}{*}{\scriptsize Avg. Accuracy} & {\scriptsize LLaMA-2-7B} & { 69.00} & {\scriptsize 64.58} & {\scriptsize 63.65} & {\scriptsize 61.17} & {\scriptsize 56.91} & {\scriptsize 47.36} \\

{\scriptsize } & {\scriptsize LLaMA-2-13B} & { 71.70} & {\scriptsize 69.18} & {\scriptsize 68.19} & {\scriptsize 67.07} & {\scriptsize 65.93} & {\scriptsize 63.76} \\

\cmidrule(lr){2-8}

{\scriptsize } & {\scriptsize LLaMA-3-8B} & { 72.57} & {\scriptsize 70.43} & {\scriptsize 68.77} & {\scriptsize 67.58} & {\scriptsize 65.79} & {\scriptsize 60.44} \\

\bottomrule

\vspace{-0.6cm}
\end{tabular}
\end{center}
\end{table}

\textbf{Effect of the SDR group size.} A key feature of our approach is the use of smaller group sizes compared to other methods, as we can perform low-precision computations without the need for dequantization, thereby maintaining hardware performance. However, smaller group sizes do result in an increased size of the cached flag data. Meanwhile, other group-based quantization approaches require additional data due to the scale factors generated for each group. For example, FP32 and FP16 scale factors add an overhead of 0.25 and 0.125 bits per value, respectively, when the group size is 128. We evaluate QRazor under various group size settings to assess its efficiency.

Table \ref{grouptable} presents the average accuracy of LLaMA models across various tasks when QRazor is applied with different group sizes in W4A4KV4 configuration. As the group size increases, capturing salient bits becomes more challenging due to the larger number of data points sharing the same razoring point. Consequently, significant accuracy degradation is observed in some tasks when tested with a group size of 128, which shares four flag bits within the group, resulting in an effective bit-width of approximately 4.03 bits per data. Based on the average accuracy drop, a group size of 32 or smaller provides minimal accuracy loss across various models, outperforming other methods. Group sizes of 16 and 32 also yield reasonable effective bit-widths of 4.25 and 4.125, respectively, which are comparable to those of group-based quantization methods with a group size of 128. This explains why group sizes of 16 and 32 are selected as the primary configurations for comparison.

\begin{table}[t]
\caption{Comparison of power and area for MAC units}
\label{hardwaretable}
\begin{center}
\setlength{\tabcolsep}{1pt}
\scriptsize
\tiny
\resizebox{8.2cm}{!}{
\begin{tabular}{c|cccc}
\toprule
\multicolumn{1}{c}{} & \multicolumn{1}{c}{$\;\;$ FP 16$\times$16 $\;\;$} & \multicolumn{1}{c}{$\;\;$ INT 16$\times$8 $\;\;$} & \multicolumn{1}{c}{$\;\;$ INT 8$\times$8 $\;\;$} & \multicolumn{1}{c}{$\;\;$ INT 4$\times$4 $\;\;$} \\ 
\multicolumn{1}{c}{} & \multicolumn{1}{c}{MAC} & \multicolumn{1}{c}{MAC} & \multicolumn{1}{c}{MAC} & \multicolumn{1}{c}{Proposed} \\ 
\midrule
\textbf{$\;\;$Area ($\mu m^2$)$\;\;$} \\
\midrule
Multiplier & 3042.2 & 1052.2 & 559.4 & 112 \\
Shifter & 0 & 0 & 0 & 156.5 \\
Reg. + Accm. & 1127.1 & 631 & 431 & 385.3 \\
\midrule
Total & 4169.3 & 1683.2 & 990.4 & 653.8 \\
\midrule\midrule
\textbf{$\;\;$Power ($mW$)$\;\;$} \\
\midrule
Multiplier & 0.3378 & 0.0506 & 0.023 & 0.0028 \\
Shifter & 0 & 0 & 0 & 0.0067 \\
Reg. + Accm. & 0.1242 & 0.0733 & 0.0581 & 0.0451 \\
\midrule
Total & 0.4620 & 0.1239 & 0.0811 & 0.0546 \\
\bottomrule
\end{tabular}}
\vspace{-0.35cm}
\end{center}
\end{table}

\subsection{Hardware Efficiency} \label {subsec:hardware}
The hardware efficiency of the processing unit design described in section \ref{subsec:arith} is evaluated by fully implementing the register-transfer-level designs in Verilog and synthesizing them with $\displaystyle Synopsys$ $\displaystyle Design$ $\displaystyle Compiler$ using an industrial LP 65nm library. The power consumption is extracted using $\displaystyle PrimeTime$ $\displaystyle PX$. As shown in Table 2, our proposed design achieves a 61.2\% reduction in area compared to the 16×8 INT MAC unit, which represents the base precision arithmetic unit, and a 34\% reduction compared to an 8×8 INT MAC unit, which serves as the standard precision for integer-based GEMM in most GPUs \citep{qserve}. In terms of power consumption, our unit achieves reductions of 56\% and 33.7\% compared to the 16$\times$8 INT MAC and 8$\times$8 INT MAC units, respectively. Compared to the FP16 MAC unit, the area and power saving become more significant.

The significant reductions in both area and power highlight the efficiency of the proposed decompression-free method. These improvements translate into several advantages for the system, including increased computational throughput, reduced hardware resource consumption, and enhanced energy efficiency. This makes the system particularly suitable for deployment in resource-constrained environments, such as edge devices, where minimizing area and power is critical for maintaining performance.

\section{Conclusion}
We introduced QRazor, a post-training quantization (PTQ) method that enables reliable 4-bit quantization for large language models (LLMs) through two key stages: quantization and compression. Our approach effectively balances accuracy and efficiency by first quantizing weights, activations, and KV caches to 8-bit and 16-bit integer formats, followed by compressing them using the significant data razoring (SDR) technique, which preserves salient bits for low-bit computations. Experimental results on various LLMs demonstrate QRazor's reliable performance compared to state-of-the-art methods in scenarios such as W4A4 and W4A4KV4. Additionally, our hardware-optimized, decompression-free arithmetic unit significantly reduces area and power consumption, making QRazor ideal for deployment in resource-constrained environments.




\bibliography{example_paper}

\begin{thebibliography}{39}
\providecommand{\natexlab}[1]{#1}
\providecommand{\url}[1]{\texttt{#1}}
\expandafter\ifx\csname urlstyle\endcsname\relax
  \providecommand{\doi}[1]{doi: #1}\else
  \providecommand{\doi}{doi: \begingroup \urlstyle{rm}\Url}\fi

\bibitem[Ashkboos et~al.(2023)Ashkboos, Markov, Frantar, Zhong, Wang, Ren, Hoefler, and Alistarh]{quik}
Ashkboos, S., Markov, I., Frantar, E., Zhong, T., Wang, X., Ren, J., Hoefler, T., and Alistarh, D.
\newblock Quik: Towards end-to-end 4-bit inference on generative large language models, 2023.
\newblock URL \url{https://arxiv.org/abs/2310.09259}.

\bibitem[Ashkboos et~al.(2024)Ashkboos, Mohtashami, Croci, Li, Jaggi, Alistarh, Hoefler, and Hensman]{quarot}
Ashkboos, S., Mohtashami, A., Croci, M.~L., Li, B., Jaggi, M., Alistarh, D., Hoefler, T., and Hensman, J.
\newblock Quarot: Outlier-free 4-bit inference in rotated llms, 2024.
\newblock URL \url{https://arxiv.org/abs/2404.00456}.

\bibitem[Banner et~al.(2019)Banner, Nahshan, Hoffer, and Soudry]{ptq}
Banner, R., Nahshan, Y., Hoffer, E., and Soudry, D.
\newblock Post-training 4-bit quantization of convolution networks for rapid-deployment, 2019.
\newblock URL \url{https://arxiv.org/abs/1810.05723}.

\bibitem[Behdin et~al.(2023)Behdin, Acharya, Gupta, Song, Zhu, Keerthi, and Mazumder]{quantease}
Behdin, K., Acharya, A., Gupta, A., Song, Q., Zhu, S., Keerthi, S., and Mazumder, R.
\newblock Quantease: Optimization-based quantization for language models, 2023.
\newblock URL \url{https://arxiv.org/abs/2309.01885}.

\bibitem[Chee et~al.(2024)Chee, Cai, Kuleshov, and Sa]{qulp}
Chee, J., Cai, Y., Kuleshov, V., and Sa, C.~D.
\newblock Quip: 2-bit quantization of large language models with guarantees, 2024.
\newblock URL \url{https://arxiv.org/abs/2307.13304}.

\bibitem[Cheng et~al.(2024)Cheng, Zhang, Shen, Cai, He, Lv, and Liu]{signround}
Cheng, W., Zhang, W., Shen, H., Cai, Y., He, X., Lv, K., and Liu, Y.
\newblock Optimize weight rounding via signed gradient descent for the quantization of llms, 2024.
\newblock URL \url{https://arxiv.org/abs/2309.05516}.

\bibitem[Dai et~al.(2021)Dai, Venkatesan, Ren, Zimmer, Dally, and Khailany]{vsquant}
Dai, S., Venkatesan, R., Ren, H., Zimmer, B., Dally, W.~J., and Khailany, B.
\newblock Vs-quant: Per-vector scaled quantization for accurate low-precision neural network inference, 2021.
\newblock URL \url{https://arxiv.org/abs/2102.04503}.

\bibitem[de~Amorim et~al.(2023)de~Amorim, Cavalcanti, and Cruz]{scaling}
de~Amorim, L.~B., Cavalcanti, G.~D., and Cruz, R.~M.
\newblock The choice of scaling technique matters for classification performance.
\newblock \emph{Applied Soft Computing}, 133:\penalty0 109924, January 2023.
\newblock ISSN 1568-4946.
\newblock \doi{10.1016/j.asoc.2022.109924}.
\newblock URL \url{http://dx.doi.org/10.1016/j.asoc.2022.109924}.

\bibitem[Dettmers et~al.(2022)Dettmers, Lewis, Belkada, and Zettlemoyer]{LLM.int8}
Dettmers, T., Lewis, M., Belkada, Y., and Zettlemoyer, L.
\newblock Gpt3.int8(): 8-bit matrix multiplication for transformers at scale.
\newblock In Koyejo, S., Mohamed, S., Agarwal, A., Belgrave, D., Cho, K., and Oh, A. (eds.), \emph{Advances in Neural Information Processing Systems}, volume~35, pp.\  30318--30332. Curran Associates, Inc., 2022.
\newblock URL \url{https://proceedings.neurips.cc/paper_files/paper/2022/file/c3ba4962c05c49636d4c6206a97e9c8a-Paper-Conference.pdf}.

\bibitem[Dettmers et~al.(2023)Dettmers, Svirschevski, Egiazarian, Kuznedelev, Frantar, Ashkboos, Borzunov, Hoefler, and Alistarh]{spqr}
Dettmers, T., Svirschevski, R., Egiazarian, V., Kuznedelev, D., Frantar, E., Ashkboos, S., Borzunov, A., Hoefler, T., and Alistarh, D.
\newblock Spqr: A sparse-quantized representation for near-lossless llm weight compression, 2023.
\newblock URL \url{https://arxiv.org/abs/2306.03078}.

\bibitem[Frantar et~al.(2023)Frantar, Ashkboos, Hoefler, and Alistarh]{gptq}
Frantar, E., Ashkboos, S., Hoefler, T., and Alistarh, D.
\newblock Gptq: Accurate post-training quantization for generative pre-trained transformers, 2023.
\newblock URL \url{https://arxiv.org/abs/2210.17323}.

\bibitem[Gao et~al.(2023)Gao, Tow, Abbasi, Biderman, Black, DiPofi, Foster, Golding, Hsu, Le~Noac'h, Li, McDonell, Muennighoff, Ociepa, Phang, Reynolds, Schoelkopf, Skowron, Sutawika, Tang, Thite, Wang, Wang, and Zou]{eval-harness}
Gao, L., Tow, J., Abbasi, B., Biderman, S., Black, S., DiPofi, A., Foster, C., Golding, L., Hsu, J., Le~Noac'h, A., Li, H., McDonell, K., Muennighoff, N., Ociepa, C., Phang, J., Reynolds, L., Schoelkopf, H., Skowron, A., Sutawika, L., Tang, E., Thite, A., Wang, B., Wang, K., and Zou, A.
\newblock A framework for few-shot language model evaluation, 12 2023.
\newblock URL \url{https://zenodo.org/records/10256836}.

\bibitem[Grattafiori et~al.(2024)Grattafiori, Dubey, Jauhri, Pandey, Kadian, Al-Dahle, Letman, Mathur, Schelten, Vaughan, Yang, Fan, Goyal, Hartshorn, Yang, Mitra, Sravankumar, Korenev, Hinsvark, Rao, Zhang, Rodriguez, Gregerson, Spataru, Roziere, Biron, Tang, Chern, Caucheteux, Nayak, Bi, Marra, McConnell, Keller, Touret, Wu, Wong, Ferrer, Nikolaidis, Allonsius, Song, Pintz, Livshits, Wyatt, Esiobu, Choudhary, Mahajan, Garcia-Olano, Perino, Hupkes, Lakomkin, AlBadawy, Lobanova, Dinan, Smith, Radenovic, Guzmán, Zhang, Synnaeve, Lee, Anderson, Thattai, Nail, Mialon, Pang, Cucurell, Nguyen, Korevaar, Xu, Touvron, Zarov, Ibarra, Kloumann, Misra, Evtimov, Zhang, Copet, Lee, Geffert, Vranes, Park, Mahadeokar, Shah, van~der Linde, Billock, Hong, Lee, Fu, Chi, Huang, Liu, Wang, Yu, Bitton, Spisak, Park, Rocca, Johnstun, Saxe, Jia, Alwala, Prasad, Upasani, Plawiak, Li, Heafield, Stone, El-Arini, Iyer, Malik, Chiu, Bhalla, Lakhotia, Rantala-Yeary, van~der Maaten, Chen, Tan, Jenkins, Martin, Madaan, Malo, Blecher,
  Landzaat, de~Oliveira, Muzzi, Pasupuleti, Singh, Paluri, Kardas, Tsimpoukelli, Oldham, Rita, Pavlova, Kambadur, Lewis, Si, Singh, Hassan, Goyal, Torabi, Bashlykov, Bogoychev, Chatterji, Zhang, Duchenne, Çelebi, Alrassy, Zhang, Li, Vasic, Weng, Bhargava, Dubal, Krishnan, Koura, Xu, He, Dong, Srinivasan, Ganapathy, Calderer, Cabral, Stojnic, Raileanu, Maheswari, Girdhar, Patel, Sauvestre, Polidoro, Sumbaly, Taylor, Silva, Hou, Wang, Hosseini, Chennabasappa, Singh, Bell, Kim, Edunov, Nie, Narang, Raparthy, Shen, Wan, Bhosale, Zhang, Vandenhende, Batra, Whitman, Sootla, Collot, Gururangan, Borodinsky, Herman, Fowler, Sheasha, Georgiou, Scialom, Speckbacher, Mihaylov, Xiao, Karn, Goswami, Gupta, Ramanathan, Kerkez, Gonguet, Do, Vogeti, Albiero, Petrovic, Chu, Xiong, Fu, Meers, Martinet, Wang, Wang, Tan, Xia, Xie, Jia, Wang, Goldschlag, Gaur, Babaei, Wen, Song, Zhang, Li, Mao, Coudert, Yan, Chen, Papakipos, Singh, Srivastava, Jain, Kelsey, Shajnfeld, Gangidi, Victoria, Goldstand, Menon, Sharma, Boesenberg,
  Baevski, Feinstein, Kallet, Sangani, Teo, Yunus, Lupu, Alvarado, Caples, Gu, Ho, Poulton, Ryan, Ramchandani, Dong, Franco, Goyal, Saraf, Chowdhury, Gabriel, Bharambe, Eisenman, Yazdan, James, Maurer, Leonhardi, Huang, Loyd, Paola, Paranjape, Liu, Wu, Ni, Hancock, Wasti, Spence, Stojkovic, Gamido, Montalvo, Parker, Burton, Mejia, Liu, Wang, Kim, Zhou, Hu, Chu, Cai, Tindal, Feichtenhofer, Gao, Civin, Beaty, Kreymer, Li, Adkins, Xu, Testuggine, David, Parikh, Liskovich, Foss, Wang, Le, Holland, Dowling, Jamil, Montgomery, Presani, Hahn, Wood, Le, Brinkman, Arcaute, Dunbar, Smothers, Sun, Kreuk, Tian, Kokkinos, Ozgenel, Caggioni, Kanayet, Seide, Florez, Schwarz, Badeer, Swee, Halpern, Herman, Sizov, Guangyi, Zhang, Lakshminarayanan, Inan, Shojanazeri, Zou, Wang, Zha, Habeeb, Rudolph, Suk, Aspegren, Goldman, Zhan, Damlaj, Molybog, Tufanov, Leontiadis, Veliche, Gat, Weissman, Geboski, Kohli, Lam, Asher, Gaya, Marcus, Tang, Chan, Zhen, Reizenstein, Teboul, Zhong, Jin, Yang, Cummings, Carvill, Shepard, McPhie,
  Torres, Ginsburg, Wang, Wu, U, Saxena, Khandelwal, Zand, Matosich, Veeraraghavan, Michelena, Li, Jagadeesh, Huang, Chawla, Huang, Chen, Garg, A, Silva, Bell, Zhang, Guo, Yu, Moshkovich, Wehrstedt, Khabsa, Avalani, Bhatt, Mankus, Hasson, Lennie, Reso, Groshev, Naumov, Lathi, Keneally, Liu, Seltzer, Valko, Restrepo, Patel, Vyatskov, Samvelyan, Clark, Macey, Wang, Hermoso, Metanat, Rastegari, Bansal, Santhanam, Parks, White, Bawa, Singhal, Egebo, Usunier, Mehta, Laptev, Dong, Cheng, Chernoguz, Hart, Salpekar, Kalinli, Kent, Parekh, Saab, Balaji, Rittner, Bontrager, Roux, Dollar, Zvyagina, Ratanchandani, Yuvraj, Liang, Alao, Rodriguez, Ayub, Murthy, Nayani, Mitra, Parthasarathy, Li, Hogan, Battey, Wang, Howes, Rinott, Mehta, Siby, Bondu, Datta, Chugh, Hunt, Dhillon, Sidorov, Pan, Mahajan, Verma, Yamamoto, Ramaswamy, Lindsay, Lindsay, Feng, Lin, Zha, Patil, Shankar, Zhang, Zhang, Wang, Agarwal, Sajuyigbe, Chintala, Max, Chen, Kehoe, Satterfield, Govindaprasad, Gupta, Deng, Cho, Virk, Subramanian, Choudhury,
  Goldman, Remez, Glaser, Best, Koehler, Robinson, Li, Zhang, Matthews, Chou, Shaked, Vontimitta, Ajayi, Montanez, Mohan, Kumar, Mangla, Ionescu, Poenaru, Mihailescu, Ivanov, Li, Wang, Jiang, Bouaziz, Constable, Tang, Wu, Wang, Wu, Gao, Kleinman, Chen, Hu, Jia, Qi, Li, Zhang, Zhang, Adi, Nam, Yu, Wang, Zhao, Hao, Qian, Li, He, Rait, DeVito, Rosnbrick, Wen, Yang, Zhao, and Ma]{llama3}
Grattafiori, A., Dubey, A., Jauhri, A., Pandey, A., Kadian, A., Al-Dahle, A., Letman, A., Mathur, A., Schelten, A., Vaughan, A., Yang, A., Fan, A., Goyal, A., Hartshorn, A., Yang, A., Mitra, A., Sravankumar, A., Korenev, A., Hinsvark, A., Rao, A., Zhang, A., Rodriguez, A., Gregerson, A., Spataru, A., Roziere, B., Biron, B., Tang, B., Chern, B., Caucheteux, C., Nayak, C., Bi, C., Marra, C., McConnell, C., Keller, C., Touret, C., Wu, C., Wong, C., Ferrer, C.~C., Nikolaidis, C., Allonsius, D., Song, D., Pintz, D., Livshits, D., Wyatt, D., Esiobu, D., Choudhary, D., Mahajan, D., Garcia-Olano, D., Perino, D., Hupkes, D., Lakomkin, E., AlBadawy, E., Lobanova, E., Dinan, E., Smith, E.~M., Radenovic, F., Guzmán, F., Zhang, F., Synnaeve, G., Lee, G., Anderson, G.~L., Thattai, G., Nail, G., Mialon, G., Pang, G., Cucurell, G., Nguyen, H., Korevaar, H., Xu, H., Touvron, H., Zarov, I., Ibarra, I.~A., Kloumann, I., Misra, I., Evtimov, I., Zhang, J., Copet, J., Lee, J., Geffert, J., Vranes, J., Park, J., Mahadeokar, J.,
  Shah, J., van~der Linde, J., Billock, J., Hong, J., Lee, J., Fu, J., Chi, J., Huang, J., Liu, J., Wang, J., Yu, J., Bitton, J., Spisak, J., Park, J., Rocca, J., Johnstun, J., Saxe, J., Jia, J., Alwala, K.~V., Prasad, K., Upasani, K., Plawiak, K., Li, K., Heafield, K., Stone, K., El-Arini, K., Iyer, K., Malik, K., Chiu, K., Bhalla, K., Lakhotia, K., Rantala-Yeary, L., van~der Maaten, L., Chen, L., Tan, L., Jenkins, L., Martin, L., Madaan, L., Malo, L., Blecher, L., Landzaat, L., de~Oliveira, L., Muzzi, M., Pasupuleti, M., Singh, M., Paluri, M., Kardas, M., Tsimpoukelli, M., Oldham, M., Rita, M., Pavlova, M., Kambadur, M., Lewis, M., Si, M., Singh, M.~K., Hassan, M., Goyal, N., Torabi, N., Bashlykov, N., Bogoychev, N., Chatterji, N., Zhang, N., Duchenne, O., Çelebi, O., Alrassy, P., Zhang, P., Li, P., Vasic, P., Weng, P., Bhargava, P., Dubal, P., Krishnan, P., Koura, P.~S., Xu, P., He, Q., Dong, Q., Srinivasan, R., Ganapathy, R., Calderer, R., Cabral, R.~S., Stojnic, R., Raileanu, R., Maheswari, R., Girdhar,
  R., Patel, R., Sauvestre, R., Polidoro, R., Sumbaly, R., Taylor, R., Silva, R., Hou, R., Wang, R., Hosseini, S., Chennabasappa, S., Singh, S., Bell, S., Kim, S.~S., Edunov, S., Nie, S., Narang, S., Raparthy, S., Shen, S., Wan, S., Bhosale, S., Zhang, S., Vandenhende, S., Batra, S., Whitman, S., Sootla, S., Collot, S., Gururangan, S., Borodinsky, S., Herman, T., Fowler, T., Sheasha, T., Georgiou, T., Scialom, T., Speckbacher, T., Mihaylov, T., Xiao, T., Karn, U., Goswami, V., Gupta, V., Ramanathan, V., Kerkez, V., Gonguet, V., Do, V., Vogeti, V., Albiero, V., Petrovic, V., Chu, W., Xiong, W., Fu, W., Meers, W., Martinet, X., Wang, X., Wang, X., Tan, X.~E., Xia, X., Xie, X., Jia, X., Wang, X., Goldschlag, Y., Gaur, Y., Babaei, Y., Wen, Y., Song, Y., Zhang, Y., Li, Y., Mao, Y., Coudert, Z.~D., Yan, Z., Chen, Z., Papakipos, Z., Singh, A., Srivastava, A., Jain, A., Kelsey, A., Shajnfeld, A., Gangidi, A., Victoria, A., Goldstand, A., Menon, A., Sharma, A., Boesenberg, A., Baevski, A., Feinstein, A., Kallet, A.,
  Sangani, A., Teo, A., Yunus, A., Lupu, A., Alvarado, A., Caples, A., Gu, A., Ho, A., Poulton, A., Ryan, A., Ramchandani, A., Dong, A., Franco, A., Goyal, A., Saraf, A., Chowdhury, A., Gabriel, A., Bharambe, A., Eisenman, A., Yazdan, A., James, B., Maurer, B., Leonhardi, B., Huang, B., Loyd, B., Paola, B.~D., Paranjape, B., Liu, B., Wu, B., Ni, B., Hancock, B., Wasti, B., Spence, B., Stojkovic, B., Gamido, B., Montalvo, B., Parker, C., Burton, C., Mejia, C., Liu, C., Wang, C., Kim, C., Zhou, C., Hu, C., Chu, C.-H., Cai, C., Tindal, C., Feichtenhofer, C., Gao, C., Civin, D., Beaty, D., Kreymer, D., Li, D., Adkins, D., Xu, D., Testuggine, D., David, D., Parikh, D., Liskovich, D., Foss, D., Wang, D., Le, D., Holland, D., Dowling, E., Jamil, E., Montgomery, E., Presani, E., Hahn, E., Wood, E., Le, E.-T., Brinkman, E., Arcaute, E., Dunbar, E., Smothers, E., Sun, F., Kreuk, F., Tian, F., Kokkinos, F., Ozgenel, F., Caggioni, F., Kanayet, F., Seide, F., Florez, G.~M., Schwarz, G., Badeer, G., Swee, G., Halpern, G.,
  Herman, G., Sizov, G., Guangyi, Zhang, Lakshminarayanan, G., Inan, H., Shojanazeri, H., Zou, H., Wang, H., Zha, H., Habeeb, H., Rudolph, H., Suk, H., Aspegren, H., Goldman, H., Zhan, H., Damlaj, I., Molybog, I., Tufanov, I., Leontiadis, I., Veliche, I.-E., Gat, I., Weissman, J., Geboski, J., Kohli, J., Lam, J., Asher, J., Gaya, J.-B., Marcus, J., Tang, J., Chan, J., Zhen, J., Reizenstein, J., Teboul, J., Zhong, J., Jin, J., Yang, J., Cummings, J., Carvill, J., Shepard, J., McPhie, J., Torres, J., Ginsburg, J., Wang, J., Wu, K., U, K.~H., Saxena, K., Khandelwal, K., Zand, K., Matosich, K., Veeraraghavan, K., Michelena, K., Li, K., Jagadeesh, K., Huang, K., Chawla, K., Huang, K., Chen, L., Garg, L., A, L., Silva, L., Bell, L., Zhang, L., Guo, L., Yu, L., Moshkovich, L., Wehrstedt, L., Khabsa, M., Avalani, M., Bhatt, M., Mankus, M., Hasson, M., Lennie, M., Reso, M., Groshev, M., Naumov, M., Lathi, M., Keneally, M., Liu, M., Seltzer, M.~L., Valko, M., Restrepo, M., Patel, M., Vyatskov, M., Samvelyan, M., Clark,
  M., Macey, M., Wang, M., Hermoso, M.~J., Metanat, M., Rastegari, M., Bansal, M., Santhanam, N., Parks, N., White, N., Bawa, N., Singhal, N., Egebo, N., Usunier, N., Mehta, N., Laptev, N.~P., Dong, N., Cheng, N., Chernoguz, O., Hart, O., Salpekar, O., Kalinli, O., Kent, P., Parekh, P., Saab, P., Balaji, P., Rittner, P., Bontrager, P., Roux, P., Dollar, P., Zvyagina, P., Ratanchandani, P., Yuvraj, P., Liang, Q., Alao, R., Rodriguez, R., Ayub, R., Murthy, R., Nayani, R., Mitra, R., Parthasarathy, R., Li, R., Hogan, R., Battey, R., Wang, R., Howes, R., Rinott, R., Mehta, S., Siby, S., Bondu, S.~J., Datta, S., Chugh, S., Hunt, S., Dhillon, S., Sidorov, S., Pan, S., Mahajan, S., Verma, S., Yamamoto, S., Ramaswamy, S., Lindsay, S., Lindsay, S., Feng, S., Lin, S., Zha, S.~C., Patil, S., Shankar, S., Zhang, S., Zhang, S., Wang, S., Agarwal, S., Sajuyigbe, S., Chintala, S., Max, S., Chen, S., Kehoe, S., Satterfield, S., Govindaprasad, S., Gupta, S., Deng, S., Cho, S., Virk, S., Subramanian, S., Choudhury, S.,
  Goldman, S., Remez, T., Glaser, T., Best, T., Koehler, T., Robinson, T., Li, T., Zhang, T., Matthews, T., Chou, T., Shaked, T., Vontimitta, V., Ajayi, V., Montanez, V., Mohan, V., Kumar, V.~S., Mangla, V., Ionescu, V., Poenaru, V., Mihailescu, V.~T., Ivanov, V., Li, W., Wang, W., Jiang, W., Bouaziz, W., Constable, W., Tang, X., Wu, X., Wang, X., Wu, X., Gao, X., Kleinman, Y., Chen, Y., Hu, Y., Jia, Y., Qi, Y., Li, Y., Zhang, Y., Zhang, Y., Adi, Y., Nam, Y., Yu, Wang, Zhao, Y., Hao, Y., Qian, Y., Li, Y., He, Y., Rait, Z., DeVito, Z., Rosnbrick, Z., Wen, Z., Yang, Z., Zhao, Z., and Ma, Z.
\newblock The llama 3 herd of models, 2024.
\newblock URL \url{https://arxiv.org/abs/2407.21783}.

\bibitem[Guo et~al.(2023)Guo, Tang, Hu, Leng, Zhang, Yang, Liu, Guo, and Zhu]{olive}
Guo, C., Tang, J., Hu, W., Leng, J., Zhang, C., Yang, F., Liu, Y., Guo, M., and Zhu, Y.
\newblock Olive: Accelerating large language models via hardware-friendly outlier-victim pair quantization.
\newblock In \emph{Proceedings of the 50th Annual International Symposium on Computer Architecture}, volume~36, pp.\  1–15. ACM, June 2023.
\newblock \doi{10.1145/3579371.3589038}.
\newblock URL \url{http://dx.doi.org/10.1145/3579371.3589038}.

\bibitem[Ho et~al.(2020)Ho, Le, and Chang]{o2a}
Ho, N.-D., Le, M.-S., and Chang, I.-J.
\newblock O-2a: Low overhead dnn compression with outlier-aware approximation.
\newblock In \emph{2020 57th ACM/IEEE Design Automation Conference (DAC)}, pp.\  1--6, 2020.
\newblock \doi{10.1109/DAC18072.2020.9218594}.

\bibitem[Hooper et~al.(2024)Hooper, Kim, Mohammadzadeh, Mahoney, Shao, Keutzer, and Gholami]{kvquant}
Hooper, C., Kim, S., Mohammadzadeh, H., Mahoney, M.~W., Shao, Y.~S., Keutzer, K., and Gholami, A.
\newblock Kvquant: Towards 10 million context length llm inference with kv cache quantization, 2024.
\newblock URL \url{https://arxiv.org/abs/2401.18079}.

\bibitem[Kim et~al.(2024)Kim, Hooper, Gholami, Dong, Li, Shen, Mahoney, and Keutzer]{squeezellm}
Kim, S., Hooper, C., Gholami, A., Dong, Z., Li, X., Shen, S., Mahoney, M.~W., and Keutzer, K.
\newblock Squeezellm: Dense-and-sparse quantization, 2024.
\newblock URL \url{https://arxiv.org/abs/2306.07629}.

\bibitem[Latotzke et~al.(2022)Latotzke, Balim, and Gemmeke]{absmax}
Latotzke, C., Balim, B., and Gemmeke, T.
\newblock Post-training quantization for energy efficient realization of deep neural networks, 2022.
\newblock URL \url{https://arxiv.org/abs/2210.07906}.

\bibitem[Lee et~al.(2024)Lee, Jin, Kim, Kim, and Park]{owq}
Lee, C., Jin, J., Kim, T., Kim, H., and Park, E.
\newblock Owq: Outlier-aware weight quantization for efficient fine-tuning and inference of large language models, 2024.
\newblock URL \url{https://arxiv.org/abs/2306.02272}.

\bibitem[Li et~al.(2023{\natexlab{a}})Li, Li, Zhang, and Chu]{normtweaking}
Li, L., Li, Q., Zhang, B., and Chu, X.
\newblock Norm tweaking: High-performance low-bit quantization of large language models, 2023{\natexlab{a}}.
\newblock URL \url{https://arxiv.org/abs/2309.02784}.

\bibitem[Li et~al.(2023{\natexlab{b}})Li, Zhang, Li, Yao, Zhang, Chu, Sun, Du, and Xie]{fptq}
Li, Q., Zhang, Y., Li, L., Yao, P., Zhang, B., Chu, X., Sun, Y., Du, L., and Xie, Y.
\newblock Fptq: Fine-grained post-training quantization for large language models, 2023{\natexlab{b}}.
\newblock URL \url{https://arxiv.org/abs/2308.15987}.

\bibitem[Lin et~al.(2024{\natexlab{a}})Lin, Tang, Tang, Yang, Chen, Wang, Xiao, Dang, Gan, and Han]{AWQ}
Lin, J., Tang, J., Tang, H., Yang, S., Chen, W.-M., Wang, W.-C., Xiao, G., Dang, X., Gan, C., and Han, S.
\newblock Awq: Activation-aware weight quantization for llm compression and acceleration, 2024{\natexlab{a}}.
\newblock URL \url{https://arxiv.org/abs/2306.00978}.

\bibitem[Lin et~al.(2024{\natexlab{b}})Lin, Tang, Yang, Zhang, Xiao, Gan, and Han]{qserve}
Lin, Y., Tang, H., Yang, S., Zhang, Z., Xiao, G., Gan, C., and Han, S.
\newblock Qserve: W4a8kv4 quantization and system co-design for efficient llm serving, 2024{\natexlab{b}}.
\newblock URL \url{https://arxiv.org/abs/2405.04532}.

\bibitem[Liu et~al.(2024{\natexlab{a}})Liu, Gong, Wei, Dong, Cai, and Zhuang]{qllm}
Liu, J., Gong, R., Wei, X., Dong, Z., Cai, J., and Zhuang, B.
\newblock Qllm: Accurate and efficient low-bitwidth quantization for large language models, 2024{\natexlab{a}}.
\newblock URL \url{https://arxiv.org/abs/2310.08041}.

\bibitem[Liu et~al.(2023)Liu, Oguz, Zhao, Chang, Stock, Mehdad, Shi, Krishnamoorthi, and Chandra]{qat}
Liu, Z., Oguz, B., Zhao, C., Chang, E., Stock, P., Mehdad, Y., Shi, Y., Krishnamoorthi, R., and Chandra, V.
\newblock Llm-qat: Data-free quantization aware training for large language models, 2023.
\newblock URL \url{https://arxiv.org/abs/2305.17888}.

\bibitem[Liu et~al.(2024{\natexlab{b}})Liu, Zhao, Fedorov, Soran, Choudhary, Krishnamoorthi, Chandra, Tian, and Blankevoort]{Spinquant}
Liu, Z., Zhao, C., Fedorov, I., Soran, B., Choudhary, D., Krishnamoorthi, R., Chandra, V., Tian, Y., and Blankevoort, T.
\newblock Spinquant: Llm quantization with learned rotations, 2024{\natexlab{b}}.
\newblock URL \url{https://arxiv.org/abs/2405.16406}.

\bibitem[Paperno et~al.(2016)Paperno, Kruszewski, Lazaridou, Pham, Bernardi, Pezzelle, Baroni, Boleda, and Fernández]{lambada}
Paperno, D., Kruszewski, G., Lazaridou, A., Pham, Q.~N., Bernardi, R., Pezzelle, S., Baroni, M., Boleda, G., and Fernández, R.
\newblock The lambada dataset: Word prediction requiring a broad discourse context, 2016.
\newblock URL \url{https://arxiv.org/abs/1606.06031}.

\bibitem[Patro \& Sahu(2015)Patro and Sahu]{normalization}
Patro, S. G.~K. and Sahu, K.~K.
\newblock Normalization: A preprocessing stage, 2015.
\newblock URL \url{https://arxiv.org/abs/1503.06462}.

\bibitem[Shao et~al.(2024)Shao, Chen, Zhang, Xu, Zhao, Li, Zhang, Gao, Qiao, and Luo]{omniquant}
Shao, W., Chen, M., Zhang, Z., Xu, P., Zhao, L., Li, Z., Zhang, K., Gao, P., Qiao, Y., and Luo, P.
\newblock Omniquant: Omnidirectionally calibrated quantization for large language models, 2024.
\newblock URL \url{https://arxiv.org/abs/2308.13137}.

\bibitem[Sheng et~al.(2023)Sheng, Zheng, Yuan, Li, Ryabinin, Fu, Xie, Chen, Barrett, Gonzalez, Liang, Ré, Stoica, and Zhang]{flexgen}
Sheng, Y., Zheng, L., Yuan, B., Li, Z., Ryabinin, M., Fu, D.~Y., Xie, Z., Chen, B., Barrett, C., Gonzalez, J.~E., Liang, P., Ré, C., Stoica, I., and Zhang, C.
\newblock Flexgen: High-throughput generative inference of large language models with a single gpu, 2023.
\newblock URL \url{https://arxiv.org/abs/2303.06865}.

\bibitem[Steinerberger(2024)]{hadamard}
Steinerberger, S.
\newblock A note on approximate hadamard matrices, 2024.
\newblock URL \url{https://arxiv.org/abs/2402.13202}.

\bibitem[Touvron et~al.(2023)Touvron, Martin, Stone, Albert, Almahairi, Babaei, Bashlykov, Batra, Bhargava, Bhosale, Bikel, Blecher, Ferrer, Chen, Cucurull, Esiobu, Fernandes, Fu, Fu, Fuller, Gao, Goswami, Goyal, Hartshorn, Hosseini, Hou, Inan, Kardas, Kerkez, Khabsa, Kloumann, Korenev, Koura, Lachaux, Lavril, Lee, Liskovich, Lu, Mao, Martinet, Mihaylov, Mishra, Molybog, Nie, Poulton, Reizenstein, Rungta, Saladi, Schelten, Silva, Smith, Subramanian, Tan, Tang, Taylor, Williams, Kuan, Xu, Yan, Zarov, Zhang, Fan, Kambadur, Narang, Rodriguez, Stojnic, Edunov, and Scialom]{llama2}
Touvron, H., Martin, L., Stone, K., Albert, P., Almahairi, A., Babaei, Y., Bashlykov, N., Batra, S., Bhargava, P., Bhosale, S., Bikel, D., Blecher, L., Ferrer, C.~C., Chen, M., Cucurull, G., Esiobu, D., Fernandes, J., Fu, J., Fu, W., Fuller, B., Gao, C., Goswami, V., Goyal, N., Hartshorn, A., Hosseini, S., Hou, R., Inan, H., Kardas, M., Kerkez, V., Khabsa, M., Kloumann, I., Korenev, A., Koura, P.~S., Lachaux, M.-A., Lavril, T., Lee, J., Liskovich, D., Lu, Y., Mao, Y., Martinet, X., Mihaylov, T., Mishra, P., Molybog, I., Nie, Y., Poulton, A., Reizenstein, J., Rungta, R., Saladi, K., Schelten, A., Silva, R., Smith, E.~M., Subramanian, R., Tan, X.~E., Tang, B., Taylor, R., Williams, A., Kuan, J.~X., Xu, P., Yan, Z., Zarov, I., Zhang, Y., Fan, A., Kambadur, M., Narang, S., Rodriguez, A., Stojnic, R., Edunov, S., and Scialom, T.
\newblock Llama 2: Open foundation and fine-tuned chat models, 2023.
\newblock URL \url{https://arxiv.org/abs/2307.09288}.

\bibitem[Wang et~al.(2024)Wang, Yin, Sun, Qi, Wang, Zhuang, Yang, and Liao]{outliertune}
Wang, J., Yin, Y., Sun, H., Qi, Q., Wang, J., Zhuang, Z., Yang, T., and Liao, J.
\newblock Outliertune: Efficient channel-wise quantization for large language models, 2024.
\newblock URL \url{https://arxiv.org/abs/2406.18832}.

\bibitem[Wei et~al.(2023)Wei, Zhang, Li, Zhang, Gong, Guo, and Liu]{outliersuppression}
Wei, X., Zhang, Y., Li, Y., Zhang, X., Gong, R., Guo, J., and Liu, X.
\newblock Outlier suppression+: Accurate quantization of large language models by equivalent and optimal shifting and scaling, 2023.
\newblock URL \url{https://arxiv.org/abs/2304.09145}.

\bibitem[Xiao et~al.(2024)Xiao, Lin, Seznec, Wu, Demouth, and Han]{smoothquant}
Xiao, G., Lin, J., Seznec, M., Wu, H., Demouth, J., and Han, S.
\newblock Smoothquant: Accurate and efficient post-training quantization for large language models, 2024.
\newblock URL \url{https://arxiv.org/abs/2211.10438}.

\bibitem[Yao et~al.(2022)Yao, Yazdani~Aminabadi, Zhang, Wu, Li, and He]{Zeroquant}
Yao, Z., Yazdani~Aminabadi, R., Zhang, M., Wu, X., Li, C., and He, Y.
\newblock Zeroquant: Efficient and affordable post-training quantization for large-scale transformers.
\newblock In Koyejo, S., Mohamed, S., Agarwal, A., Belgrave, D., Cho, K., and Oh, A. (eds.), \emph{Advances in Neural Information Processing Systems}, volume~35, pp.\  27168--27183. Curran Associates, Inc., 2022.
\newblock URL \url{https://proceedings.neurips.cc/paper_files/paper/2022/file/adf7fa39d65e2983d724ff7da57f00ac-Paper-Conference.pdf}.

\bibitem[Yuan et~al.(2023)Yuan, Niu, Liu, Liu, Wang, Shang, Sun, Wu, Wu, and Wu]{rptq}
Yuan, Z., Niu, L., Liu, J., Liu, W., Wang, X., Shang, Y., Sun, G., Wu, Q., Wu, J., and Wu, B.
\newblock Rptq: Reorder-based post-training quantization for large language models, 2023.
\newblock URL \url{https://arxiv.org/abs/2304.01089}.

\bibitem[Zhang et~al.(2023)Zhang, Zhao, Cao, Wang, Cao, Yang, Yang, Zhang, and Xu]{mofq}
Zhang, Y., Zhao, L., Cao, S., Wang, W., Cao, T., Yang, F., Yang, M., Zhang, S., and Xu, N.
\newblock Integer or floating point? new outlooks for low-bit quantization on large language models, 2023.
\newblock URL \url{https://arxiv.org/abs/2305.12356}.

\bibitem[Zhao et~al.(2024)Zhao, Lin, Zhu, Ye, Chen, Zheng, Ceze, Krishnamurthy, Chen, and Kasikci]{atom}
Zhao, Y., Lin, C.-Y., Zhu, K., Ye, Z., Chen, L., Zheng, S., Ceze, L., Krishnamurthy, A., Chen, T., and Kasikci, B.
\newblock Atom: Low-bit quantization for efficient and accurate llm serving, 2024.
\newblock URL \url{https://arxiv.org/abs/2310.19102}.

\end{thebibliography}
\bibliographystyle{icml2025}

\newpage
\appendix
\onecolumn
\section{Appendix}
\label{appendix1}

\subsection{Importance of the Weight Quantization}

To assess the impact of weight and activation compression on accuracy using the QRazor method, we conducted the following experiments. From the integer data quantized into our base precision of W8A16KV8, we analyzed the accuracy of W4A8, W8A8, and W4A16 configurations to examine the impact of weight compression in comparison to activation compression with group size of 8. As shown in Table \ref{w4w8table}, W8A8 demonstrates the highest accuracy among the three cases. The results suggest that compressing weight values into fewer bits is as sensitive as capturing activation outliers. When outlier characteristics are well-preserved, reducing activation bits using our SDR scheme has minimal impact on overall accuracy. In contrast, although weights are less affected by outliers, reducing their bit-width to 4 bits can introduce quantization errors, significantly impacting overall accuracy.
Consequently, further optimization of weight quantization can enhance the accuracy of our 4-bit LLM, such as by applying the GPTQ \citep{gptq} technique to weight values alongside our QRazor scheme, as briefly mentioned in Section \ref{subsec:ablation}.

\begin{table}[h]
\caption{Zero-shot accuracy comparison of QRazor between W4A8, W8A8, and W4A16}
\label{w4w8table}
\begin{center}
\scriptsize
\resizebox{\textwidth}{!}{ }
\begin{tabular}{ccccccccc}
\toprule
\multicolumn{3}{c}{} & \multicolumn{5}{c}{\textbf{Zero-shot Accuracy} $\uparrow$} \\  
\cmidrule(lr){4-8} 
\multicolumn{1}{c}{Model} & \multicolumn{1}{c}{\#Bits} & \multicolumn{1}{c}{Method} & 
\multicolumn{1}{c}{PIQA} & \multicolumn{1}{c}{ARC-e} & \multicolumn{1}{c}{ARC-c} & \multicolumn{1}{c}{HellaSwag} & \multicolumn{1}{c}{Winogrande}\\

\midrule


\multirow{4}{*}{LLaMA-2-7B} & { FP16} & { Baseline} & { 79.13} & { 74.39} & { 45.97} & { 76.21} & { 69.30} \\

\cmidrule(lr){2-8}

{  } & { W4A8} & { } & { 77.04} & { 75.09} & { 45.48} & { 74.87} & { 68.90}\\

{  } & { W8A8} & {QRazor} & { 77.75} & { 75.79} & { 45.82} & { 75.34} & { 69.14} \\

{  } & { W4A16} & { } & { 77.20} & { 74.91} & { 44.82} & { 74.99} & { 68.98} \\

\midrule

\multirow{4}{*}{ LLaMA-2-13B} & { FP16} & { Baseline} & { 80.23} & { 77.82} & { 48.76} & { 79.39} & { 72.30} \\

\cmidrule(lr){2-8}

{  } & { W4A8} & { } & { 78.56} & { 78.60} & { 44.82} & { 78.33} & { 70.88} \\

{  } & { W8A8} & {QRazor} & { 78.67} & { 78.77} & { 45.15} & { 79.37} & { 72.38}\\

{  } & { W4A16} & { } & { 78.56} & { 78.77} & { 45.15} & { 78.51} & { 70.92} \\

\bottomrule
\end{tabular}
\end{center}
\end{table}

\subsection{SDR Encoding: Detecting the Razoring Point with Bitwise OR Operations} \label{appendix3}

\begin{figure}[h]
\begin{center}
\fbox{
    \begin{minipage}{0.97\columnwidth}
        \includegraphics[width=\columnwidth]{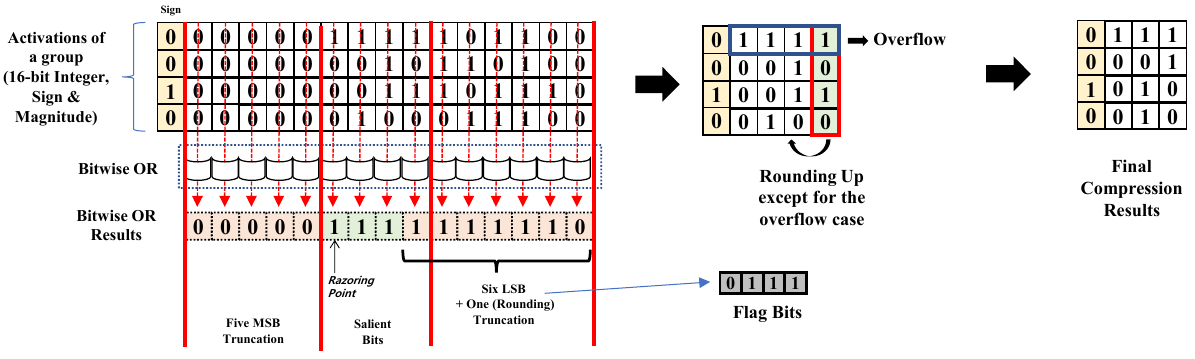}
    \end{minipage}
}
\end{center}
\vspace{-0.3cm}
\caption{SDR Encoding Scheme Consisting of Bitwise OR Operations for detecting the razoring point.}
\label{fig:bitwiseor}

\end{figure}

As mentioned in Section \ref{subsec:Stage2}, the primary concept of our SDR scheme is its streamlined encoding and decoding process compared to other methods, offering the key advantage of on-the-fly compression and decompression, particularly for activations. We have discussed the details of our decompression-free arithmetic design, which facilitates end-to-end arithmetic operations while bypassing the decompression process. Although we have outlined the overall encoding concept of our SDR scheme, we have not yet provided detailed information regarding the hardware aspects. Figure \ref{fig:bitwiseor} illustrates the entire SDR process, starting with detecting the razoring point, which can be implemented as simple parallel bitwise OR operations within a group. The result of the bitwise OR operation effectively identifies the MSB position within the group, and the flag bits are automatically generated based on this position. The nearest-to-round method, combined with truncation, preserves the salient bits at the target precision.

\subsection{Perplexity}
We evaluated the performance of our QRazor scheme using additional perplexity metric of \texttt{Lambada-OpenAI} dataset \citep{lambada} for W4A4, W4A8, W4A4KV4, and W4A8KV4 configurations on the LLaMA-2-7B and LLaMA-2-13B models. For calibration, 128 randomly selected samples from wikitext2 data has been used. 

\begin{table}[h]
\caption{Perplexity of different group size with Lambada-OpenAI}
\label{perplextable}
\begin{center}
\scriptsize
\resizebox{\textwidth}{!}{ }
\begin{tabular}{ccccccccccc}
\toprule
\multicolumn{3}{c}{} & \multicolumn{6}{c}{\textbf{Group Size Perplexity} $\downarrow$} \\  
\cmidrule(lr){4-9} 
\multicolumn{1}{c}{Model} & \multicolumn{1}{c}{\#Bits} & \multicolumn{1}{c}{Method} & 
\multicolumn{1}{c}{Baseline} &\multicolumn{1}{c}{8} & \multicolumn{1}{c}{16} & \multicolumn{1}{c}{32} & \multicolumn{1}{c}{64} & \multicolumn{1}{c}{128}\\

\midrule

\multirow{4}{*}{ LLaMA-2-7B} & { W4A8} & \multirow{4}{*}{ QRazor} & \multirow{4}{*}{ 3.40} & { 3.83} & { 4.01} & { 4.34} & { 4.52} & { 4.65}\\

{  } & { W4A4} & { } & { } & { 4.24} & { 4.47} & { 5.04} & { 5.83} & { 7.94}\\

{  } & { W4A8KV4} & { } & { } & { 4.14} & { 4.31} & { 4.63} & { 5.04} & { 6.11}\\

{  } & { W4A4KV4} & { } & { } & { 4.29} & { 4.98} & { 5.91} & { 7.87} & { 19.23}\\

\midrule

\multirow{4}{*}{ LLaMA-2-13B} & { W4A8} & \multirow{4}{*}{ QRazor} & \multirow{4}{*}{ 3.04} & { 3.36} & {3.28} & { 3.34} & { 3.38} & { 3.39}\\

{  } & { W4A4} & { } & { } & { 3.45} & { 3.64} & { 3.93} & { 4.58} & { 6.22}\\

{  } & { W4A8KV4} & { } & { } & { 3.36} & { 3.45} & { 3.54} & { 3.63} & { 4.15}\\

{  } & { W4A4KV4} & { } & { } & { 3.59} & { 3.79} & { 4.27} & { 5.09} & { 9.35}\\

\bottomrule
\end{tabular}
\end{center}
\end{table} 

\vspace{1cm}

\subsection{FLOPs \& OPs Comparison Results}
To evaluate the computational efficiency of our proposed quantization scheme relative to the approach outlined in the QuaRot paper, we conducted a comparative analysis of the FLOPs and IOPs involved in the attention layer of a transformer architecture. QuaRot employs two primary operations—Hadamard product and rotation operation—for outlier-aware quantization. These operations effectively reduce the impact of outliers during quantization by transforming datasets into lower-variance distributions. This process enables efficient mapping of data to 4-bit representations using conventional per-tensor or per-channel quantization techniques.

However, the Hadamard product and rotation operations introduce substantial computational overhead during both quantization and dequantization. Notably, the dequantization process in QuaRot incurs a significant number of additional FLOPs, as it involves dequantization computations for weights as well. Consequently, while QuaRot is effective for quantization, its reliance on FLOPs-intensive operations reduces computational efficiency when compared to other methods with similar IOPs.

In contrast, our proposed quantization scheme simplifies the dequantization process by utilizing shifting operations immediately after INT matrix-multiplication(matmul) to perform decompression. 

\begin{table}[h]
\caption{OPs comparison table of rotation and SDR computations}
\label{Operation}
\centering
\renewcommand{\arraystretch}{1.5}
\resizebox{\textwidth}{!}{
\begin{tabular}{|c|c|c|c|c|}
\hline
\textbf{Metric} & \multicolumn{2}{c|}{\textbf{Hadamard (Rotation)}} & \multicolumn{2}{c|}{\textbf{QRazor (SDR)}} \\ \hline
\textbf{Operation}& \textbf{Single Hadamard} & \textbf{Hadamard Heads} & \textbf{SDR Compression} & \textbf{Barrel Shifter} \\ \hline
\textbf{Formula} & $M \times N$ & $H \times M \times N$ & ($M \times N \times 2$)/$G$ & ($M \times N$)/$G$ \\ \hline
\textbf{Result ($M = 128, N = 64, H = 8, G= 32$)} 
& $8,192$ FLOPs & $65,536$ FLOPs & $512$ IOPs & $256$ IOPs \\ \hline
\end{tabular}
}
\end{table}

\clearpage



\subsection{Overall Accuracy Results}
We present the overall accuracy results experimented by QRazor method with LLaMA2 models in Table \ref{overalltable}, encompassing all cases across different bit-widths and group sizes. Notably, W4A8 demonstrates negligible accuracy in all scenarios, including W4A8KV4. In W4A4 and W4A4KV4 configurations, group sizes of 16 and 32 exhibit reliable accuracy across all cases.

\begin{table}[h]
\caption{Zero-shot accuracy of quantized LLaMA2 models on five common sense tasks.}
\label{overalltable}
\begin{center}
\scriptsize
\begin{tabular}{cccccccccc}
\toprule
\multicolumn{3}{c}{} & \multicolumn{6}{c}{\textbf{Zero-shot Accuracy} $\uparrow$} \\  
\cmidrule(lr){4-8} 
\multicolumn{1}{c}{Model} & \multicolumn{1}{c}{\#Bits} & \multicolumn{1}{c}{Group Size} & 
\multicolumn{1}{c}{PIQA} & \multicolumn{1}{c}{ARC-e} & \multicolumn{1}{c}{ARC-c}  & \multicolumn{1}{c}{HellaSwag} & \multicolumn{1}{c}{Winogrande} & \multicolumn{1}{c}{Avg.} \\

\midrule

\multirow{24}{*}{\scriptsize LLaMA-2-7B } & {\scriptsize FP16} & {\scriptsize Baseline} & {\scriptsize 79.13} & {\scriptsize 74.39} & {\scriptsize 45.97} & {\scriptsize 76.21} & {\scriptsize 69.30} & {\scriptsize 69.00} \\

\cmidrule(lr){2-9} 

{\scriptsize } & {\scriptsize } & {\scriptsize 8} & {\scriptsize 77.09} & {\scriptsize 73.44} & {\scriptsize 44.53} & {\scriptsize 74.97} & {\scriptsize 68.98} & {\scriptsize 67.80} \\
{\scriptsize } & {\scriptsize } & {\scriptsize 16} & {\scriptsize 77.48} & {\scriptsize 74.39} & {\scriptsize 40.13} & {\scriptsize 74.87} & {\scriptsize 68.35} & {\scriptsize 67.04} \\
{\scriptsize } & {\scriptsize W4A8} & {\scriptsize 32} & {\scriptsize 77.09} & {\scriptsize 74.91} & {\scriptsize 39.14} & {\scriptsize 74.47} & {\scriptsize 68.98} & {\scriptsize 66.92} \\
{\scriptsize } & {\scriptsize } & {\scriptsize 64} & {\scriptsize 76.71} & {\scriptsize 74.26} & {\scriptsize 43.82} & {\scriptsize 73.96} & {\scriptsize 67.25} & {\scriptsize 67.20} \\
{\scriptsize } & {\scriptsize } & {\scriptsize 128} & {\scriptsize 76.17} & {\scriptsize 73.09} & {\scriptsize 43.11} & {\scriptsize 72.76} & {\scriptsize 67.88} & {\scriptsize 66.60} \\

\cmidrule(lr){2-9} 

{\scriptsize } & {\scriptsize } & {\scriptsize 8} & {\scriptsize 75.30} & {\scriptsize 72.88} & {\scriptsize 41.14} & {\scriptsize 73.40} & {\scriptsize 66.30} & {\scriptsize 65.80} \\
{\scriptsize } & {\scriptsize } & {\scriptsize 16} & {\scriptsize 75.84} & {\scriptsize 72.63} & {\scriptsize 42.47} & {\scriptsize 72.96} & {\scriptsize 65.67} & {\scriptsize 65.91} \\
{\scriptsize } & {\scriptsize W4A4} & {\scriptsize 32} & {\scriptsize 73.67} & {\scriptsize 70.70} & {\scriptsize 39.46} & {\scriptsize 71.46} & {\scriptsize 64.09} & {\scriptsize 63.88} \\
{\scriptsize } & {\scriptsize } & {\scriptsize 64} & {\scriptsize 73.99} & {\scriptsize 69.12} & {\scriptsize 39.46} & {\scriptsize 69.06} & {\scriptsize 62.98} & {\scriptsize 62.92} \\
{\scriptsize } & {\scriptsize } & {\scriptsize 128} & {\scriptsize 69.91} & {\scriptsize 63.16} & {\scriptsize 32.78}& {\scriptsize 63.31} & {\scriptsize 59.27} & {\scriptsize 57.69} \\

\cmidrule(lr){2-9} 

{\scriptsize } & {\scriptsize } & {\scriptsize 8} & {\scriptsize 76.55} & {\scriptsize 73.86} & {\scriptsize 43.14} & {\scriptsize 73.81} & {\scriptsize 68.59} & {\scriptsize 67.19} \\
{\scriptsize } & {\scriptsize } & {\scriptsize 16} & {\scriptsize 76.66} & {\scriptsize 71.75} & {\scriptsize 43.82} & {\scriptsize 72.97} & {\scriptsize 68.11} & {\scriptsize 66.66} \\
{\scriptsize } & {\scriptsize W4A8KV4} & {\scriptsize 32} & {\scriptsize 75.46} & {\scriptsize 71.58} & {\scriptsize 43.38} & {\scriptsize 66.95} & {\scriptsize 62.04} & {\scriptsize 63.88} \\
{\scriptsize } & {\scriptsize } & {\scriptsize 64} & {\scriptsize 69.97} & {\scriptsize 61.58} & {\scriptsize 34.78} & {\scriptsize 61.32} & {\scriptsize 56.91} & {\scriptsize 56.91} \\
{\scriptsize } & {\scriptsize } & {\scriptsize 128} & {\scriptsize 61.81} & {\scriptsize 52.63} & {\scriptsize 27.76} & {\scriptsize 43.07} & {\scriptsize 51.54} & {\scriptsize 47.36} \\

\cmidrule(lr){2-9} 

{\scriptsize } & {\scriptsize } & {\scriptsize 8} & {\scriptsize 75.35} & {\scriptsize 70.70} & {\scriptsize 39.80} & {\scriptsize 71.63} & {\scriptsize 65.43} & {\scriptsize 64.58} \\
{\scriptsize } & {\scriptsize } & {\scriptsize 16} & {\scriptsize 73.39} & {\scriptsize 70.88} & {\scriptsize 39.80} & {\scriptsize 70.15} & {\scriptsize 64.01} & {\scriptsize 63.65} \\
{\scriptsize } & {\scriptsize W4A4KV4} & {\scriptsize 32} & {\scriptsize 73.23} & {\scriptsize 67.54} & {\scriptsize 37.46} & {\scriptsize 67.16} & {\scriptsize 60.46} & {\scriptsize 61.17} \\
{\scriptsize } & {\scriptsize } & {\scriptsize 64} & {\scriptsize 69.97} & {\scriptsize 61.58} & {\scriptsize 34.78} & {\scriptsize 61.32} & {\scriptsize 56.91} & {\scriptsize 56.91} \\
{\scriptsize } & {\scriptsize } & {\scriptsize 128} & {\scriptsize 61.81} & {\scriptsize 52.63} & {\scriptsize 27.76} & {\scriptsize 43.07} & {\scriptsize 51.54} & {\scriptsize 47.36} \\

\midrule

\multirow{24}{*}{\scriptsize LLaMA-2-13B} & {\scriptsize FP16} & {\scriptsize Baseline} & {\scriptsize 80.23} & {\scriptsize 77.82} & {\scriptsize 48.76} & {\scriptsize 79.39} & {\scriptsize 72.30} & {\scriptsize 71.70} \\

\cmidrule(lr){2-9} 

{\scriptsize } & {\scriptsize } & {\scriptsize 8} & {\scriptsize 78.78} & {\scriptsize 78.77} & {\scriptsize 45.82} & {\scriptsize 78.48} & {\scriptsize 71.74} & {\scriptsize 70.72} \\
{\scriptsize } & {\scriptsize } & {\scriptsize 16} & {\scriptsize 79.33} & {\scriptsize 76.42} & {\scriptsize 45.14} & {\scriptsize 78.24} & {\scriptsize 71.74} & {\scriptsize 70.17} \\
{\scriptsize } & {\scriptsize W4A8} & {\scriptsize 32} & {\scriptsize 79.00} & {\scriptsize 76.07} & {\scriptsize 46.48} & {\scriptsize 78.07} & {\scriptsize 69.93} & {\scriptsize 69.91} \\
{\scriptsize } & {\scriptsize } & {\scriptsize 64} & {\scriptsize 78.67} & {\scriptsize 75.89} & {\scriptsize 45.48} & {\scriptsize 77.97} & {\scriptsize 70.56} & {\scriptsize 69.71} \\
{\scriptsize } & {\scriptsize } & {\scriptsize 128} & {\scriptsize 78.62} & {\scriptsize 75.60} & {\scriptsize 44.14} & {\scriptsize 77.70} & {\scriptsize 69.93} & {\scriptsize 69.20} \\

\cmidrule(lr){2-9} 

{\scriptsize } & {\scriptsize } & {\scriptsize 8} & {\scriptsize 77.97} & {\scriptsize 75.19} & {\scriptsize 42.81} & {\scriptsize 77.19} & {\scriptsize 70.72} & {\scriptsize 68.78} \\
{\scriptsize } & {\scriptsize } & {\scriptsize 16} & {\scriptsize 77.48} & {\scriptsize 73.09} & {\scriptsize 43.47} & {\scriptsize 76.83} & {\scriptsize 70.09} & {\scriptsize 68.17} \\
{\scriptsize } & {\scriptsize W4A4} & {\scriptsize 32} & {\scriptsize 76.82} & {\scriptsize 71.86} & {\scriptsize 41.80} & {\scriptsize 76.30} & {\scriptsize 68.59} & {\scriptsize 67.07} \\
{\scriptsize } & {\scriptsize } & {\scriptsize 64} & {\scriptsize 76.12} & {\scriptsize 70.98} & {\scriptsize 41.36} & {\scriptsize 74.74} & {\scriptsize 67.01} & {\scriptsize 66.04} \\
{\scriptsize } & {\scriptsize } & {\scriptsize 128} & {\scriptsize 74.97} & {\scriptsize 70.11} & {\scriptsize 38.54} & {\scriptsize 73.12} & {\scriptsize 62.04} & {\scriptsize 63.76} \\

\cmidrule(lr){2-9} 

{\scriptsize } & {\scriptsize } & {\scriptsize 8} & {\scriptsize 79.05} & {\scriptsize 77.19} & {\scriptsize 49.16} & {\scriptsize 78.32} & {\scriptsize 71.43} & {\scriptsize 71.03} \\
{\scriptsize } & {\scriptsize } & {\scriptsize 16} & {\scriptsize 78.51} & {\scriptsize 76.67} & {\scriptsize 45.15} & {\scriptsize 77.92} & {\scriptsize 71.19} & {\scriptsize 69.89} \\
{\scriptsize } & {\scriptsize W4A8KV4} & {\scriptsize 32} & {\scriptsize 78.24} & {\scriptsize 76.84} & {\scriptsize 47.83} & {\scriptsize 77.54} & {\scriptsize 69.93} & {\scriptsize 70.08}\\
{\scriptsize } & {\scriptsize } & {\scriptsize 64} & {\scriptsize 78.02} & {\scriptsize 76.14} & {\scriptsize 47.81} & {\scriptsize 76.71} & {\scriptsize 69.77} & {\scriptsize 69.69} \\
{\scriptsize} & {\scriptsize } & {\scriptsize 128} & {\scriptsize 77.20} & {\scriptsize 73.33} & {\scriptsize 44.15} & {\scriptsize 74.66} & {\scriptsize 68.98} & {\scriptsize 67.67} \\

\cmidrule(lr){2-9} 

{\scriptsize } & {\scriptsize } & {\scriptsize 8} & {\scriptsize 76.77} & {\scriptsize 75.61} & {\scriptsize 46.49} & {\scriptsize 76.55} & {\scriptsize 71.11} & {\scriptsize 69.31} \\
{\scriptsize } & {\scriptsize } & {\scriptsize 16} & {\scriptsize 76.55} & {\scriptsize 75.44} & {\scriptsize 44.15} & {\scriptsize 75.38} & {\scriptsize 68.43} & {\scriptsize 67.99} \\
{\scriptsize } & {\scriptsize W4A4KV4} & {\scriptsize 32} & {\scriptsize 75.35} & {\scriptsize 71.75} & {\scriptsize 44.28} & {\scriptsize 73.61} & {\scriptsize 66.77} & {\scriptsize 66.35} \\
{\scriptsize } & {\scriptsize } & {\scriptsize 64} & {\scriptsize 74.59} & {\scriptsize 71.40} & {\scriptsize 42.12} & {\scriptsize 70.84} & {\scriptsize 62.19} & {\scriptsize 64.23} \\
{\scriptsize } & {\scriptsize } & {\scriptsize 128} & {\scriptsize 68.82} & {\scriptsize 62.81} & {\scriptsize 32.78} & {\scriptsize 58.64} & {\scriptsize 55.17} & {\scriptsize 55.64} \\

\bottomrule
\end{tabular}
\end{center}
\end{table}

\clearpage

\subsection{Accuracy Comparison with SOTA Works}
Here, we conduct accuracy comparison with 4bit quantization SOTA works with Mistral-7B model. In all configurations, our method achieved highest performance by average accuaracy drop of less than 6\% in five common sense tasks.

\begin{table}[h]
\caption{Zero-shot accuracy of Mistral-7B models on five common sense tasks.}
\label{mistraltable}
\renewcommand{\arraystretch}{0.9}
\begin{center}
\tiny
\resizebox{\textwidth}{!}{ }
\begin{tabular}{cccccccccc}
\toprule
\multicolumn{3}{c}{} & \multicolumn{6}{c}{\textbf{Zero-shot Accuracy} $\uparrow$} \\  
\cmidrule(lr){4-8} 
\multicolumn{1}{c}{Model} & \multicolumn{1}{c}{\#Bits} & \multicolumn{1}{c}{Method} & 
\multicolumn{1}{c}{PIQA} & \multicolumn{1}{c}{ARC-e} & \multicolumn{1}{c}{ARC-c} & \multicolumn{1}{c}{HellaSwag} & \multicolumn{1}{c}{Winogrande} & \multicolumn{1}{c}{Avg} \\

\midrule

\multirow{12}{*}{ Mistral-7B} & { FP16} & { Baseline} & { 80.74} & { 81.58} & { 50.16} & { 61.25} & { 74.27} & { 69.60} \\

\cmidrule(lr){2-9}

{ } & \multirow{4}{*}{ W4A4 } & { SmoothQuant} &{ 57.94} & { 35.14} & { 21.75} & { 30.51} & { 48.30} & { 38.73}  \\

{ } & { } & {OS+} & { 66.70} & { 56.73} & { 30.20} & { 42.39} & { 52.01} & { 49.61} \\

{ } & { } & { AWQ} & { 66.26} & { 54.16} & { 30.80} & { 43.35} & { 53.67} & { 49.67} \\

{ } & { } & { TesseraQ*} & { 72.19} & { 65.90} & { 33.78}  & { 49.02} & { 57.61} & { 55.71} \\

\cmidrule(lr){2-9}

{ } & { W4A4  $\displaystyle g16$ } & \multirow{2}{*}{ QRazor} &{77.75} & { 77.98} & { 45.48} & { 57.70} & { 69.77} & { 65.74}  \\

{ } & { W4A4  $\displaystyle g32$} & { } & { 77.80} & { 75.39} & { 43.47} & { 56.20} & { 68.90} & { 64.35} \\

\cmidrule(lr){2-9}

{ } & { W4A4KV4  $\displaystyle g16$} & \multirow{2}{*}{ QRazor} & { 77.78} & { 77.34} & { 44.48} & { 58.09} & { 68.56} & { 65.25} \\

{ } & { W4A4KV4  $\displaystyle g32$} & { } & { 77.53} & { 76.09} & { 42.24}  & { 56.38} & { 66.38} & { 63.72} \\

\bottomrule
\end{tabular}
\end{center}
\end{table}

\vspace{1cm}

\subsection{The Quantization Flow of QRazor in Transformers}

To illustrate the quantization flow of the QRazor method, we provide a detailed depiction of the Transformer layer. In comparison to methods like QuaRot\citep{quarot} and SpinQuant\citep{Spinquant}, our approach uniquely quantizes the Query, enabling a decompression-free INT4 matrix multiplication for the $Q \cdot K^\top$ operation within the Multi-Head Attention layer. This capability significantly reduces the computational burden across the Transformer layer and improves overall throughput.

\begin{figure}[h]
\begin{center}
\fbox{
    \begin{minipage}{0.97\columnwidth}
        \includegraphics[width=\columnwidth]{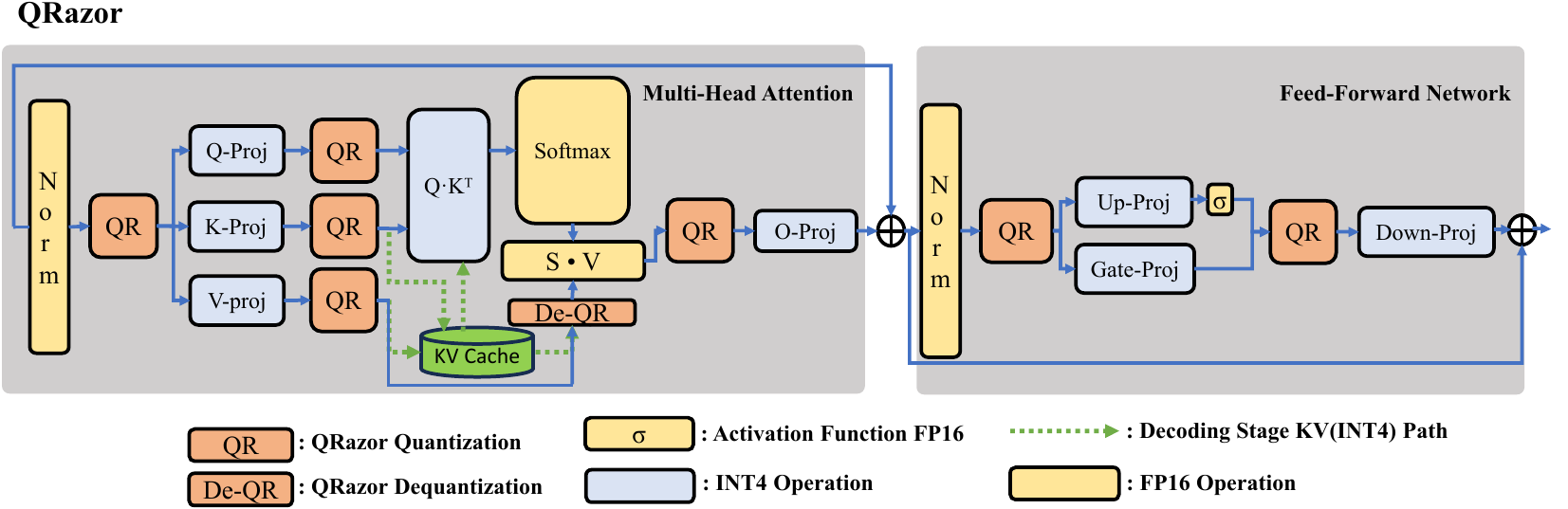}
    \end{minipage}
}
\end{center}
\vspace{-0.3cm}
\caption{The QRazor quantization flow in attention and feed-forward network layers}

\end{figure}

\end{document}